%% file: main.tex
\documentclass[10pt]{article} % For LaTeX2e
\usepackage[accepted]{tmlr}
\input{math_commands.tex}

\usepackage{hyperref}
\usepackage{url}
\usepackage{graphicx}
\usepackage{booktabs}
\usepackage{mathtools}
\usepackage{adjustbox}
\usepackage{multirow}

\title{Target-conditioned GFlowNet for Structure-based Drug Design}

\author{\name Tony Shen$^{\ast}$\textit{$^{a}$}, Seonghwan Seo\textit{$^{b}$}, Grayson Lee\textit{$^{a}$},  Mohit Pandey\textit{$^{e}$}, Jason Smith\textit{$^{e}$}, Artem Cherkasov\textit{$^{e}$}, Woo Youn Kim\textit{$^{b,c,d}$}, Martin Ester$^{\ast}$\textit{$^{a}$}
\\
\addr $^{\ast}$~Corresponding author. Contact: tony$\_$shen$\_$4@sfu.ca, ester@cs.sfu.ca \\
\addr \textit{$^{a}$}~School of Computing Science, Simon Fraser University, Burnaby, Canada. \\
\addr \textit{$^{b}$}~Department of Chemistry, KAIST, Daejeon, Republic of Korea. \\
\addr \textit{$^{c}$}~Graduate School of Data Science, KAIST, Daejeon, Republic of Korea. \\
\addr \textit{$^{d}$}~HITS Inc., Seoul, Republic of Korea. \\
\addr \textit{$^{e}$}~Vancouver Prostate Centre, University of British Columbia, Vancouver, Canada.
}

\def\month{09}  % Insert correct month for camera-ready version
\def\year{2024} % Insert correct year for camera-ready version
\def\openreview{\url{https://openreview.net/forum?id=N8cPv95zOU}} % Insert correct link to OpenReview for camera-ready version

\begin{document}

\maketitle

% fail to generate novel desirable molecules 

\begin{abstract}
Searching the vast chemical space for drug-like molecules that bind with a protein pocket is a challenging task in drug discovery. Recently, structure-based generative models have been introduced which promise to be more efficient by learning to generate molecules for any given protein structure. However, since they learn the distribution of a limited protein-ligand complex dataset, structure-based methods do not yet outperform optimization-based methods that generate binding molecules for just one pocket. To overcome limitations on data while leveraging learning across protein targets, we choose to model the reward distribution conditioned on pocket structure, instead of the training data distribution. We design \textsc{TacoGFN}, a novel GFlowNet-based approach for structure-based drug design, which can generate molecules conditioned on any protein pocket structure with probabilities proportional to its affinity and property rewards. In the generative setting for CrossDocked2020 benchmark, \textsc{TacoGFN} attains a state-of-the-art success rate of $56.0\%$ and $-8.44$ kcal/mol in median Vina Dock score while improving the generation time by multiple orders of magnitude. Fine-tuning \textsc{TacoGFN} further improves the median Vina Dock score to $-10.93$ kcal/mol and the success rate to $88.8\%$, outperforming all optimization-based methods.
\end{abstract}

\section{Introductions}
% Introduction to Structure Based Drug Discovery
Structure-based drug design (SBDD) leverages target protein structures to search for high-affinity drug molecules. Due to the growing availability of protein structures from ML protein structure prediction methods \citep{jumper2021highly}, and many novel targets identified from high-throughput perturbation experiments \citep{Replogle2022}, SBDD is becoming an increasingly powerful approach in drug discovery. It currently takes 13-15 years and between US \$2 billion and \$3 billion for a single drug to be developed and approved \citep{Pushpakom2018}. The substantial expense and time of drug development not only impose a significant burden on healthcare systems but also amplify societal risks during global health crises, such as COVID-19. There is an urgent need to expedite the design of novel drug candidates for new protein targets.

% Traditional drug discovery 
% Traditionally, molecular docking has been used to virtually screen libraries of molecules for interaction with a target protein.
Traditionally, virtual screening has been used to discover binding molecules by predicting supramolecular interactions between ligands and a target protein with molecular docking. Its efficacy is impeded by the exhaustive nature of its search, and by the high computational cost of molecular docking. To overcome this challenge, many optimization-based methods \citep{bengio2021flow, fu2022reinforced, MOODlee2023exploring,  reidenbach2024evosbdd} have been proposed to generate high-affinity molecules for one protein pocket only. These methods typically do not take protein structure as input and use docking measures as the reward. Recently, structure-based generative models \citep{luo20213d, peng2022pocket2mol, guan2023decompdiff} have been proposed to design molecules (ligands) conditioned on the pocket structure. These methods learn the distribution of a training dataset of protein-ligand complexes. Structure-based generative model has the advantage of learning generalized protein-ligand interaction patterns by leveraging different pocket structures during training.

% As the currently largest molecule library contains 38B molecules \citep{grygorenko2020enamine}, exhaustive virtual screening goes far beyond current computational capabilities. 
% Partly because molecules in virtual libraries are syntheziable from purchasable chemical building block, virtual screening is a common approach in real world drug discovery campaigns. 

% Generative model introduction 

However, due to the high cost of the experiments, the size of the training datasets for SBDD, i.e. high-quality protein-ligand binding structure data, is relatively small. PDBBind \citep{Liu2014}, the underlying dataset of the standard CrossDocked2020 benchmark \citep{Francoeur2020}, contains only 19,443 protein-ligand complexes. After removing common biomolecules (lipids, peptides, carbohydrates, and nucleotides) and duplicates, only 4,200 unique drug-like molecules remain \citep{Powers2023SBDD}. This is only a tiny fraction of the entire chemical space of drug-like molecules that is assumed to consist of $\sim 10^{60}$ molecules \citep{LIPINSKI19973}. As a consequence, existing structure-based generative models relying on data-distribution learning model only a very small part of the overall chemical space and have struggled to generate novel molecules with significantly improved properties \citep{MOODlee2023exploring}; Therefore, existing structure-based generative methods do not yet outperform optimization-based methods on a single target.

In summary, many current optimization-based methods cannot leverage learning from across protein pocket structures - they are limited to a reward function based on affinity to one single protein target per model. On the other hand, while structure-based models can learn generalized protein-ligand interaction patterns by training on different pockets, they are restricted to modelling a limited data distribution. Therefore they are unable to effectively explore the wider chemical space for desirable molecules, and under perform optimization-based methods. Our goal in this paper is to both leverage learning across protein target structures, and overcome the challenge of limited data distribution. To the best of our knowledge, TacoGFN is the first RL model to address the challenging task of modelling a family of reward functions induced from all pocket structures.

In this paper, we frame the task of structure-based molecule generation as learning a reward distribution instead of a training data distribution, and adopt GFlowNet \citet{bengio2021flow} - an energy-based generative model for generating combinatorial objects. We propose \textsc{TacoGFN}, a \textit{\underline{Ta}rget \underline{Co}nditioned \underline{G}enerative \underline{F}low \underline{N}etwork} that generates molecules conditioned on any given protein pocket structures, guided by affinity to the pocket, drug-likeness and synthesizability measures. This formulation allows us to explore the greater chemical space as we are no longer constrained to a fixed dataset. We performed an experimental evaluation on the standard CrossDocked benchmark dataset and demonstrated that \textsc{TacoGFN} clearly outperforms state-of-the-art structure-based generative methods, improving the success rate to 56.0\% from the previous best of 24.5\%. \textsc{TacoGFN} shows even stronger performance with fine-tuning (\textsc{TacoGFN+FT}), resulting in -10.93 kcal/mol in median Vina Dock score and up to 88.8\% in success rate, outperforming all optimization-based models.

To summarize, the main contributions of this paper are:
\begin{itemize}
    \item We point out the limitations of previous data-based distribution learning, and instead propose to frame structure-based molecule generation as learning from the reward distribution. To this end, we propose \textsc{TacoGFN}, the first application of GFlowNet for learning a family of molecule distributions conditioned on protein pocket structures. 

    % \item We introduce \textsc{TacoGFN+FT}, which fine-tunes the base model for a specific target pocket. 
    
    \item We introduced a novel pharmacophore-based affinity predictor, with improved generalization capabilities using a pre-trained pharmacophore representation, enabling fast affinity evaluation.
% e have
% further  where coarse-graining to the protein
% pocket is shown to achieve more accurate and robust predictions than existing architectures and protein
% representations.
    
    \item \textsc{TacoGFN} clearly outperforms the state-of-the-art structure-based methods on the CrossDocked2020 benchmark and demonstrates the benefit of reward-based distribution learning.

% We further show by leveraging the learning from generating diverse molecules for various protein pocket structures, 

   \item \textsc{TacoGFN+FT} outperforms the optimization methods focused on a single pocket only, learning from diverse molecules for various protein pocket structures.  \textsc{TacoGFN+FT} attains the best Vina Dock score, high-affinity rate and success rate among all optimization-based methods.
    % Tony: previous papers have done this too
    % \item \textcolor{blue}{We construct the fragment library via retrosynthetic rule to generate more synthesizable and chemically reliable molecules.}
    % \item We introduce a docking score prediction model with novel pharmacophore priors which is much more efficient than existing docking programs.
    % \item To our best knowledge, TacoGFN is the first method that frames the pocket-conditioned molecule generation task as an RL problem - learning a policy that generates candidates with probabilities proportional to reward based on docking score and desired properties. 
    % \item We propose modification to GFlowNet to incorporate protein pocket structure context for molecule generation. 
    % \item To solve this problem, we propose an extension of GFlowNet to incorporate protein pocket structure context for target conditioned molecule generation. 
    % \item A docking score prediction model that takes the protein and molecule structure is introduced to reward molecules with high binding affinities. 
    % \item We performed an experimental evaluation on the Cross-Docked dataset and demonstrated that \textsc{TacoGFN} generates molecules with better docking scores and properties compared to all existing state-of-the-art methods.
\end{itemize}

\begin{figure*}[!t]
  \centering
  \includegraphics[width=0.99\textwidth]{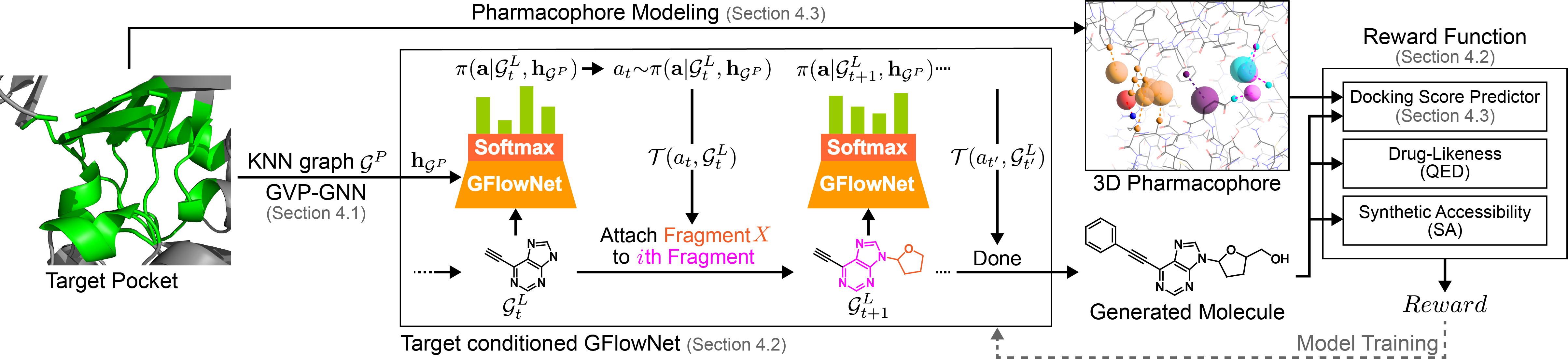}
  \caption{
    Overview of the sampling and training process of \textsc{TacoGFN}. 
  }
  \label{fig: gflownet_generation}
\end{figure*}

\section{Related Work}

\textbf{Structure-based molecular generation} aims to generate high-affinity molecules for any target pocket. They are expected to generalize on previously unseen test pockets, and therefore, do not have access to the docking oracle for the test pockets during inference or training time.  The goal for evaluating the quality of molecules generated for unseen protein structures is to measure whether SBDD models have learned generalized protein-ligand interaction patterns during training. Many approaches for this problem setting has been proposed. LiGAN \citep{masuda2020generating} uses 3D CNNs to encode the protein pocket structure and predict atom densities from the encoded latent space. 3DSBDD \citep{luo20213d} and Pocket2Mol \citep{peng2022pocket2mol} adopt an auto-regressive approach to generate molecules atom by atom. Other methods such as FLAG \citep{zhang2023molecule} and DrugGPS \citep{zhang2023drug} build molecules fragment by fragment to leverage the chemical prior. A very recent line of research employs diffusion models \citep{guan2023d, guan2023decompdiff, schneuing2023structurebased} for SBDD. TargetDiff \citep{guan2023d} is a diffusion-based method which generates atom coordinates and atom types in a non-autoregressive way, and bonds are generated in a post-processing step. DecompDiff \citep{guan2023decompdiff} is a diffusion model which generates both atoms and bonds with decomposed priors, which reflect the natural decomposition of a ligand molecule into arms and scaffold. A recent analysis paper  \citep{harris2023benchmarking} questions the assumption that explicit 3D modelling of the ligand improves performance, after finding a much higher occurrence of physical violations and fewer key interactions in molecules generated using 3D modelling. In our work, we generate molecules in 2D space to vastly reduce the search space and compute time; Our approach is shown to effectively leverage the target pocket structure for generating molecules with high affinity.

\textbf{Optimization-based molecular generation} aims to generate molecules that satisfy certain optimization goals. Compared to the distribution-based generative model, these methods can be designed to optimize for molecules with strong binding affinity to a particular protein target. Since they have access to a docking oracle during training or inference time, they are not directly comparable to the structure-based generative problem setting. Reinforcement Learning (RL) methods such as ReLeaSE \citep{ReLeaSEOlivecrona2017}, MolDQN \citep{MolDQNZhou_2019} and REINVENT \citep{REINVENTBlaschke2020} have been proposed to guide the generation of molecules toward desirable properties. MORLD \citep{Jeon2020} and MoleGuLAR \citep{MoleGuLARGoel2021} combine RL and docking calculations to design novel ligands. RGA \citep{fu2022reinforced} proposes a variant of a genetic algorithm that is guided by reinforcement learning and is pre-trained on multiple protein target structures. MOOD \citep{MOODlee2023exploring} incorporates out-of-distribution and property-guided exploration in diffusion models for goal-directed molecule generation. DecompOpt \cite{zhou2024decompopt} combines a pre-trained structure-based equivariant diffusion model with a docking-based greedy iterative optimization loop. EvoSBDD \cite{reidenbach2024evosbdd} improves efficiency by performing black-box optimization over the 1D latent space using a docking oracle.  In summary, optimization-based molecular generation evaluates the performance of the algorithm on optimizing for seen targets. Compared to the structure-based setting, which requires modelling the molecule distribution conditional to any protein pocket structure and generalizing to unseen protein pockets, optimizing the molecule distribution for a single protein pocket is a less challenging task. We refer readers to a more detailed comparison of the problem setting in Appendix \ref{appendix: compare}. 

\textbf{Protein-ligand affinity prediction.}
Predicting the affinity or the docking score of a ligand to a target, in the absence of their binding complex structure, is a difficult task. Most previous docking score prediction models have been limited to a single protein target \citep{bengio2021flow, gentile2020deepdocking}. Therefore, they are not suitable for the aim of designing high-affinity molecules for any given protein pocket structure.
% are not suitable for the context of structure-based drug design which
Recently, several methods have been proposed to predict ligand affinity for arbitrary protein targets \citep{zhang2023deepbindgcn, molecules27165114}.
However, these approaches are prone to memorizing the structural bias instead of learning the physics of protein-ligand binding and show low generalization ability to unseen ligands or proteins \citep{wallach2018most, chan2023embracing}.
To this end, we adopt a pre-trained pharmacophore representation, which only models the key interaction sites for a protein pocket. This prior improves our docking score predictor's ability to learn physical interactions and generalize to unseen data. 

\section{GFlowNet Preliminaries}
Generative Flow Networks (GFlowNets, GFN) \citep{bengio2021flow} learn a stochastic policy $\pi$ for generating a combinatorial object (such as molecular graph) $x\in \mathcal{X}$. The probability of constructing $x$, denoted as $\pi(x)$, is trained to be proportional to a non-negative reward function $R: \mathcal{X} \mapsto \mathbb{R}^+$ defined on the space $\mathcal{X}$. This property of GFlowNet is ideal for generating diverse molecules with desirable properties. Conditional GFlowNet introduced in \citep{jain2023multiobjective} simultaneously models a family of reward functions. Each conditional information, denoted as $c \in C$, induces a unique reward function $R(x|c)$. In our work, we adopt conditional GFlowNet for SBDD settings, by encoding target pocket structure as condition $c$, where $c$ is a high dimensional representation of the pocket structure.  Thus, a single GFlowNet models high-reward molecule distribution across all protein pockets. 

Each object $x$ is constructed from a sequence of actions $a \in \mathcal{A}$. In molecular settings, a molecule is constructed by inserting molecule fragments into a partially constructed fragment graph state $s \in \mathcal{S}$ \citep{bengio2021flow}. Conceptually, a GFlowNet is an acyclic graph $G = (\mathcal{S}, \mathcal{E})$, with nodes $\mathcal{S}$ and edges $\mathcal{E}$. Each transition $s\to s' \in \mathcal{E}$ via action $a \in \mathcal{A}$ corresponds to an edge in graph $G$. The transition function $\mathcal{T}: \mathcal{S}, \mathcal{A} \mapsto \mathcal{S}$ computes the new state $s' = \mathcal{T}(s, a)$ given action $a$ on state $s$. A special action moves state $s$ into terminating states $\mathcal{X} \subset \mathcal{S}$. We define the initial empty graph as $s_0$. Construction of $x$ can be defined over a trajectory of states $\tau = (s_0 {\rightarrow} s_1 {\rightarrow} \dots{\rightarrow} x)$.
%$\mathcal{T}$ is the set of all trajectories. 

Following previous studies, we introduce an exponent $\beta$ to the reward function $R(x|c)$, thus modelling $\pi(x|c, \beta) \propto R(x|c)^\beta$. This steers generating probability distribution to focus on the modes of $R(x|c)$, which is crucial for producing candidates that are high in reward. Adjusting $\beta$ allows us to manage the balance between diversity and achieving higher rewards. 

The forward transition probability $P_F(s'|s,c, \beta)$ of a GFlowNet represents the probability distribution of reaching state $s'$ from state $s$ conditioned on context $c$ and reward temperature $\beta$.  Partition function $Z(c, \beta)$ is the sum of the rewards $R(x|c)^\beta$ for all objects $x \in \mathcal{X}$ under context $c$. We adopt the Trajectory Balance objective from Equation ~\ref{eq:trajbal_objective} to efficiently learn a forward transition policy $P_F$ that generates object $x$ with probability proportional to its reward $R(x|c)^\beta$ \citep{malkin2023trajectory}. 
 
\begin{equation}
    \mathcal{L}_{TB} (\tau, c, \beta;\theta) = \left(\log \frac{Z_\theta(c, \beta) \prod_{s{\rightarrow} s' \in \tau}P_{F_\theta}(s'|s, c, \beta)}{R(x|c)^\beta}\right)^2,
    \label{eq:trajbal_objective}
\end{equation}
\( \theta \) represents the learnable parameters. Our goal is to model the probability \( \pi(x|c, \beta) \approx \frac{R(x|c)^\beta}{Z(c, \beta)} \) across all molecules \( x \in \mathcal{X} \), protein pocket contexts $c \in \mathcal{C}$ and various reward temperatures $\beta$.

\section{TacoGFN}
% Our method does not rely on the higher quality - but smaller experimental protein-ligand structure data. We only require a dataset containing the docking scores of protein-ligand pairs, which can be easily generated at scale. We trade the exploitation of a small set of known high affinity molecules for exploration of the wider chemical space.

% \textsc{TacoGFN} is a single conditioned generative model that generalizes over protein pockets.
\textsc{TacoGFN} is a structure-based molecular generative model that generalizes over all protein pockets with a single model.
By matching the reward distribution instead of limited data distribution, our method explores the greater chemical space to generate high-affinity molecules with properties desirable as a drug candidate. Furthermore, by encoding pocket structure information, \textsc{TacoGFN} and its fine-tuned variant are able to leverage learning from diverse protein pocket structures.
 % It holds the advantage of learning more efficiently over optimization-based methods that work for only one pocket.

\textbf{Problem definition.} The goal of structure-based drug design (SBDD) problem is to generate molecules with both desirable properties and strong binding affinity with respect to any given protein pocket structure. The goal for SBDD models is learning generalized protein-ligand interaction patterns during training. Therefore, SBDD models take pocket structures as input and are expected to generalize for unseen pocket structures. We refer readers to a detailed contrast of the SBDD problem setting with the optimization setting in Appendix \ref{appendix: compare}

The pocket structure is denoted as $P$ and will be represented as a $K$-nearest neighbour (KNN) residue graph. The ligand is denoted as $L$ and will be represented either as an atom graph or a fragment graph.
% The pocket structure is defined as a graph $\mathcal{G}^{pro} = (\mathcal{V}^{pro}, \mathcal{E}^{pro})$. 
% A residue node $\mathbf{v}_i^{pro} \in \mathcal{V}^{pro}$ is associated with an one-hot encoding of the residue type, and the 3D Cartesian coordinates of its backbone atoms. $\mathcal{E}^{pro}$ is the set of edges which connect pairs of adjacent residues. 
% The ligand is represented as a 2D graph $\mathcal{G}^{lig} = (\mathcal{V}^{lig}, \mathcal{E}^{lig})$. An atom node $\mathbf{v}_i^{lig} \in \mathcal{V}^{lig}$ is associated with a one-hot encoding of the atom type. An edge $\mathbf{e}_{ij}^{lig}$ represents a bond between atom $\mathbf{v}_i^{lig}$ and atom $\mathbf{v}_j^{lig} \in \mathcal{E}^{lig}$ and is associated with a one-hot encoding of the bond type. 
We define a reward function $R(L|P)$ based on a molecule's predicted docking score, drug-likeliness and synthesizability. 
Our goal is to learn a molecule generation policy that constructs molecules $L$ given protein structure $P$ with probability matching their reward (exponentiated with $\beta$), such that $\pi(L| P, \beta) \propto R(L | P)^\beta$. 

% Contrary to existing ML methods that learn the distribution of a training dataset, our RL objective is based not on a dataset but on the reward itself. 
% Since \textsc{TacoGFN}'s training objective is based on matching the reward distribution and not a fixed data distribution, it explores a much greater chemical space and discovers novel molecules beyond the known data distribution. \textcolor{blue}{Unlike optimization-based method for one pocket, \textsc{TacoGFN} also models a probability distribution conditional on protein structure. This allows our method to transfer knowledge across different pocket structures and generalize on previously unseen test pockets. 
% %In addition, by matching the generation probability to the reward, \textsc{TacoGFN} can discover molecules that improve multiple objectives compared to the training data distribution. 
% %
% %Lastly, we only require a dataset containing the docking scores of protein-ligand pairs. Unlike the high-quality near experimental protein ligand structures, docking score data can be easily generated at scale.
% In summary, \textsc{TacoGFN} effectively addresses the problem of limited data distribution experienced by existing structure-based methods; At the same time, \textsc{TacoGFN} is able leverages learning across protein target structures.}

% verage learning across
% protein target structures, and overcome the challenge of a limited data distributio
%  instead of memorizing a small set of known binding molecules.

\textbf{Method overview.}  The protein pocket is first featurized as a $K$-nearest neighbor residue graph and encoded using GVP-GNN \citep{jing2021learning} (section: \ref{section: pocket_encoding}). Then, we use this pocket embedding as the condition for GFlowNet molecule generation (section \ref{section: GFlowNet}). Finally, we reward generated molecules using our docking score prediction model which leverages protein-ligand interaction priors (section \ref{section: affinity_prediction}). 
% Optionally, a pre-trained \textsc{TacoGFN} can be later fine-tuned for individual protein targets (section \ref{section: finetune}). 
The high-level architecture of our SBDD framework is illustrated in Figure \ref{fig: gflownet_generation}.

% In this formulation, policy is trained by constructing molecules with feedback from reward function, instead of fitting on a limited dataset.

\subsection{Pocket structure encoder} \label{section: pocket_encoding}
We represent structure of the pocket $P$ as a standard $K$-nearest neighbor (KNN) residue graph $\mathcal{G}^\mathcal{P} = (\mathcal{V}^{\mathcal{P}}, \mathcal{E}^{\mathcal{P}})$ - following previous work in protein representation \citep{jing2021learning}. The $i$-th residue node $v_i^{\mathcal{P}} \in \mathcal{V}^{\mathcal{P}}$ is featurized using its geometric and chemical properties. These features include the type of residue, the dihedral angles of the atoms in the residue backbone, and the directional unit vectors.
% Specifically, the directional unit vectors are calculated from the $C_a$ atom of the residue to the $C_b$ atom, and from the $C_a$ atom to the $C_a$ atoms of adjacent residues. 
An edge $e_{ij}^{\mathcal{P}} \in \mathcal{E}^{\mathcal{P}}$ is formed if the $j$-th residue $v_j^{\mathcal{P}}$ is among the $K$-nearest neighbors of residue $v_i^{\mathcal{P}}$, as measured by the euclidean distance between their respective $C_a$ atoms.  We set the number of neighbours $K=30$. An edge is featurized with the Euclidean distance, distance along the backbone and the direction vector between the two residues. These features sufficiently describe the features of the protein pocket.

We apply a graph neural network with geometric vector perceptrons (GVP) layers \citep{jing2021learning} to the KNN pocket graph $\mathcal{G}^\mathcal{P}$  to learn the node embedding $\mathbf{h}_{v_i^{\mathcal{P}}}$ for each residue. The node embeddings \{$\mathbf{h}_{v_i^{\mathcal{P}}}\}$ are then averaged to obtain an embedding of the entire graph $\mathbf{h}_{\mathcal{G}^\mathcal{P}}$. We use GVP because it encodes the protein pocket into an embedding that is invariant to rotations and translations.

\subsection{Pocket conditioned GFlowNet} \label{section: GFlowNet}
In this section, we discuss how to employ the pocket structure, more specifically its latent embedding, to condition the GFlowNet to generate molecules that interact with a given protein pocket. Furthermore, we describe our fragment-based molecular generation framework. 

% Molecules are constructed fragment by fragment in an autoregressive manner. 

\textbf{Molecule representation.}
During molecular generation, we represent ligands as a 2D molecular graph $\mathcal{G}^L = (\mathcal{V}^L, \mathcal{E}^L)$ with node $v^L_i \in \mathcal{V}^L$ representing a molecule fragment, and directional edges $e^L_{ij} \in \mathcal{E}^L$ indicating the attachment atom of fragment $v^L_i$ that connects to fragment $v^L_j$. Since a molecule's reward (representing its desirability as a real-world drug) should be the same regardless of its predicted 3D conformation, we represent ligands as 2D graphs here. \footnote{Docking score is dependent on a molecule's 3D conformation. However, we consider the best docking score for a molecule as the reward here, which is invariant to its conformation in this context. Taking the best docking score is a valid approach, because when a compound is tested for binding in the wet lab, the compound simply binds the protein with the conformations which results in the strongest affinity. Furthermore, the predicted 3D  position of generated molecules from existing diffusion-based SBDD models often changes significantly upon re-docking and likely do not reflect the true conformation. \citep{reidenbach2024evosbdd}. }

\textbf{Fragment vocabulary construction.}
\textsc{TacoGFN} generates molecules by adding one molecular fragment at a time.
To create the vocabulary of fragments used, we extract common fragments from a chemical database in a data-driven and chemically valid way. 
% The fragment vocabulary set $\mathcal{S}^{F}$ contains fragment graphs $\mathcal{G}^{F} = (\mathcal{V}^{F}, \mathcal{E}^{F})$. Fragment graph node $v^{F}_i \in \mathcal{V}^{F}$ is associated with two features: an atom type and a flag indicating whether it can serve as an attachment point. Fragment graph edges $e^{F}_i \in \mathcal{E}^{F}$ represents a covalent bond and is associated with one feature: its bond type. 
To obtain a fragment vocabulary, we first apply BRICS decomposition \citep{Degen2008} to 250k ZINC20 \citep{irwin2020zinc20} molecules.
% BRICS breaks molecules into synthetically accessible building blocks, i.e. molecules that are easy to synthesize.
BRICS breaks molecules via retrosynthetic rules and provides synthetically accessible building blocks, i.e. molecules that are easy to prepare \citep{seo2023molecular}.
To reduce the fragment vocabulary size, we further break all single bonds connecting a heavy atom to a ring structure.
Next, we retain the fragments that occur in more than 50 (or 0.02\%) of the molecules in our ZINC set.
% Finally, we merge identical fragment graphs.
% Attachment points are atoms with a broken bond resulting from the molecule decomposition. 
Only atoms (of a fragment) in the bonds decomposed by the BRICS can join to form new fragment connections during generation time; This reduces the occurrence of non-synthesizable attachment points forming bonds.
This results in a set of 475 building block-like fragments.
As our fragments are mined from a synthetically accessible virtual library and poor synthetic accessibility is penalized in the reward function, \textsc{TacoGFN} achieves superior synthetic accessibility compared to all existing SBDD methods.

\textbf{Molecular generation framework.}
We formulate molecular generation as a sequential decision process and implement it using a GFlowNet. At the $t$-th step, the forward action policy $P_{a}$ samples an action $a$ depending on the current molecule state $\mathcal{G}^L_t$ and pocket embedding $\mathbf{h}_{\mathcal{G}^\mathcal{P}}$. The transition function $\mathcal{T}$ is a deterministic function which applies the action to the molecular graph at step $t$ to produce a molecule graph at step $t+1$.
\begin{align} 
    a_t &\sim P_{a}(\mathbf{a}|\mathcal{G}^L_t, \mathbf{h}_{\mathcal{G}^\mathcal{P}}) \\
    \mathcal{G}^L_{t+1} &= \mathcal{T}(a_t, \mathcal{G}^L_t)
\end{align}
Previous autoregressive models often formulate action prediction as a supervised task, where the goal is to predict the correct ground truth actions obtained from masked molecules \citep{peng2022pocket2mol}. Instead, our molecule generation policy $P_{a}$ aims to generate molecules with probability $P(\mathcal{G}^L|\mathbf{h}_{\mathcal{G}^\mathcal{P}})$ proportional to the reward.

\textbf{Pocket conditioned molecular generation.} \label{section: policy} Here, we adopt the architecture for molecular generation introduced in Multi-Objective GFlowNet (MO-GFN) \citep{jain2023multiobjective}. Instead of conditioning the GFlowNet on multi-objective preference, we condition the GFlowNet on pocket embeddings to learn a family of molecular distributions corresponding to a family of reward functions induced from the pocket structure diversity. 

We use a graph transformer \citep{yun2020graph} to model the probability $P_a(\mathbf{a}|\mathcal{G}^L_t, \mathbf{h}_{\mathcal{G}^\mathcal{P}})$, by taking the partially constructed molecular graph at the $t$-th time step $\mathcal{G}^L_t$ and the pocket embedding $\mathbf{h}_{\mathcal{G}^\mathcal{P}}$ as input. The input feature $\mathbf{h}^{L(0)}_i$ of molecular node $v^L_i$ is a one-hot encoding of the node's fragment type.
 % $$  with a linear transformation
The molecular edge input feature $\mathbf{e}_{ij}^{L(0)}$ is a one-hot mapping of the attachment atom index in node $v^L_i$ which connects to node $v^L_j$.
We add an additional virtual conditioning node $\mathbf{h}^{V(0)}$, featurized using pocket embedding $\mathbf{h}_{\mathcal{G}^\mathcal{P}}$, to the graph $\mathcal{G}^L_t$ \citep{pham2017graph}. This virtual node is connected to all other nodes and serves as a graph-level node to provide pocket information. After $N$ Transformer layers, the set of final node embeddings $\{\mathbf{h}^{L(N)}_i\}$ and edge embeddings $\{\mathbf{e}^{L(N)}_{ij}\}$ are obtained. The final graph level embedding $\mathbf{g}^{L(N)}$ is obtained via the concatenation of the global average pooling of the node embeddings and the final virtual node embedding $\mathbf{h}^{V(N)}$, as seen in Equation \ref{eq: transformer pooling}.
\begin{equation}
    \mathbf{g}^{L(N)} = Concat\left(AvgPool\left(\left\{\mathbf{h}^{L(N)}_i\right\}\right), \mathbf{h}^{V(N)}\right)
    \label{eq: transformer pooling}
\end{equation}

Using these final molecular graph embeddings, we follow the actions defined in previous works on fragment-based molecular generation \citep{bengio2021flow, DBLP:journals/corr/abs-1802-04364}. Full details can be found in Appendix \ref{appendix: molecular generation action}.

% 1) is a node action that connects the current node to a new node via a new edge. An MLP is applied on the node embeddings $\mathbf{h}^{L(N)}_i$ to produce logits over the fragment vocabulary for every node in the molecular graph. 2) \textit{Attachment specification} is an edge action that specifies the attachment atom in a node (fragment) which will form a single bond with the other fragment. An MLP is applied on each edge embedding $\mathbf{e}^{L(N)}_{ij}$ to produce logits over attachment point choices. 

% Generation of 2D molecule by fragments is extremely fast and vastly reduces the search space compared to 3D generation. In addition, it is shown to satisfy the objective of generating high-affinity molecules conditioned on the pocket structure.  

\textbf{Reward function.} \label{section: reward_module} A drug candidate must not only have a high affinity to the pocket but also satisfy drug-like properties and synthetizability requirements to be selected for experimental validation. We design a reward function by multiplying the normalized score from all three aspects, using QED \citep{Bickerton2012} as a measure of drug-likeliness, SA \citep{Ertl2009} as a measure of ease-of-synthesizability, and predicted docking score as a measure of affinity between ligand and protein\footnote{Computing the reward is very fast, for example, computing it for a batch of 64 protein-ligand pairs took under 0.15 seconds.}.  In drug discovery, it is crucial to design molecules that simultaneously satisfy all of these relevant properties, despite the inherent trade-offs among these properties. By multiplying the scores, the reward function ensures that a low score in any one of the factors (QED, SA, or DS) will significantly reduce the overall reward. Therefore, multiplying the scores is a more appropriate choice than summing the scores here.

We just have to optimize QED and SA up to a certain threshold for a molecule to be suitable as a drug candidate - optimizing them further does not bring additional utility \citep{COLEY2021133}. Therefore, we clip the reward for the QED or SA component to 1 when they achieve their respective threshold $t_{QED}$ and $t_{SA}$. For example, the reward function will prioritize more on optimizing QED if $t_{QED}$ is higher, at the expense of other properties such as affinity (See details of reward function in Appendix \ref{appendix: Reward function}). 

% which trains our policy to generate molecules which simultaneously satisfy all three requirements. Here $QED$ is a measure for drug-likeliness, and $SA$ measures ease-of-syntheticability. Docking score $DS$ is measures affinity between ligand and protein. To this end, we propose a fast docking score predictior (section \ref{section: affinity_prediction}). 

\begin{figure}[t]
  \centering
  \includegraphics[width=0.92\textwidth]{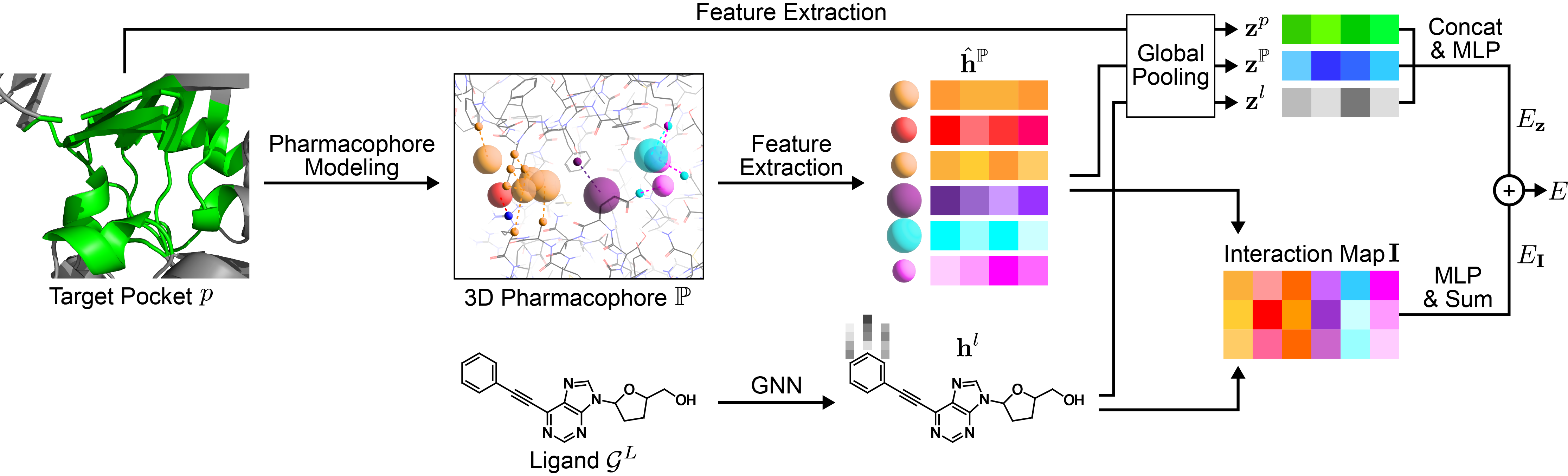}
  \caption{
    Model architecture of the docking score predictor. Each pharmacophore point is represented as a sphere and corresponds to a desired ligand characteristic for a binding interaction.
  }
  \label{fig: docking_proxy}
\end{figure}

\subsection{Docking score predictor with pharmacophore prior} \label{section: affinity_prediction}

The exploration of chemical space with a GFlowNet requires evaluating binding affinities for millions of molecules sampled during training. 
However, using molecular docking for evaluation is computationally expensive.
Here, we propose a fast ML-based docking score predictor that generalizes well across molecule and protein pocket distributions, as described in Figure \ref{fig: docking_proxy} and Equation \ref{eq: our_proxy}. (See details in Appendix \ref{appendix: docking_score_predictor})

% A pharmacophore model is a 3D point cloud, where each
% point describes the desirable motif properties (such as being
% aromatic) a ligand should possess at this geometric posi-
% tion to form energetically favourable interactions (e.g. π–π
% stacking) with the protein target.

%We use a pharmacophore $\mathbb{P}$ to represent the pocket $p$. 
% \footnote{
% A 3D Pharmacophore is a spatial pattern of pharmacophoric features that a ligand should have to ensure optimal binding to a pocket 
% }
%\footnote{A pharmacophoric feature is a spatial point at which a ligand biological trigger (e.g. aromatic ring) should be located in order to form energetically or entropically favorable interactions (e.g. $\pi$--$\pi$ stacking) with the protein target.}
We leverage PharmacoNet \citep{seo2023pharmaconet}, a recent deep learning method to obtain the pharmacophore $\mathbb{P}$\footnote{ Pharmacophore is a point set, where each point describes the desirable motif properties (such as being aromatic) a ligand should possess at this geometric position to form energetically or entropically favourable interactions (e.g. $\pi$--$\pi$ stacking) with the protein target. \citep{iupac, yang2010pharmacophore}.}  from a pocket structure $P$.To enrich the representation, we obtained the embedding of the pocket $\mathbf{z}^P$ and the embeddings of pharmacophore points $\{\hat{\mathbf{h}}^\mathbb{P}_i\}$ from PharmacoNet.
The 1D representation vectors of the pharmacophore $\mathbf{z}^\mathbb{P}$ is computed by the  global pooling of $\{\hat{\mathbf{h}}^\mathbb{P}_i\}$ .
% We leverage PharmacoNet \citep{seo2023pharmaconet},  a recent deep learning method to obtain the pharmacophore $\mathbb{P}$ from a pocket structure $\mathcal{P}$. \footnote{ Pharmacophore is a point set, where each point describes the desirable motif properties (such as being aromatic) a ligand should possess at this geometric position to form energetically favourable interactions (e.g. $\pi$--$\pi$ stacking) with the protein target. \citep{iupac, yang2010pharmacophore}.} To enrich the representation, we obtained the 3D image embedding of the pocket $\hat{\mathbf{h}}^p$ and the embedding of each pharmacophore point $\hat{\mathbf{h}}^\mathbb{P}_i$ from PharmacoNet.
% $\mathbf{z}^p$ and $\mathbf{z}^\mathbb{P}$, the 1D representation vectors of the pocket and the pharmacophore, respectively are obtained by global pooling.

During docking score prediction, we represent a ligand as an atom-level 2D molecular graph. We apply Graph Isomorphism Network \citep{hu2019strategies} to the atom graph to obtain the node embeddings $\{\mathbf{h}^L_j\}$.
$\mathbf{z}^L$ is a 1D representation vector of the ligand, obtained by the global pooling of node embeddings $\{\mathbf{h}^L_j\}$.
Then, we incorporate a pairwise interaction map $\mathbf{I}$ - computed from the outer product of $\{\hat{\mathbf{h}}^\mathbb{P}_i\}$ and $\{\mathbf{h}^L_j\}$.
% Then, we incorporate a pairwise interaction map $\mathbf{I}$, where $\mathbf{I}_{ij}$ is computed from the dot product of the embeddings of the $i$-th pharmacophore point $\mathbf{h}^\mathbb{P}_i$ with the $j$-th ligand atom $\mathbf{h}^l_j$.
% a pharmacophore point embedding and a ligand node embedding $\hat{\mathbf{h}}^\mathbb{P}_i \odot \mathbf{h}^l_j$.
This preserves the structural topology and binding interaction details.
% Notably, our docking score predictor uses the pharmacophore points involved in binding instead of atoms or amino acids to obtain the interaction map.
% This simplifies the docking score prediction task by predicting the probability of formation and energy contribution of non-covalent interactions, rather than modelling all atom-atom interactions.
Notably, our docking score predictor uses the pharmacophore points involved in binding instead of atoms or amino acids to obtain the interaction map.
It improves the generalization ability at a reduced computational cost through coarse-grained modelling of the pocket at the pharmacophore level.
\begin{equation}
    % E_\mathbf{z} &= \phi_{z} (Concat(\mathbf{z}^p, \mathbf{z}^\mathbb{P}, \mathbf{z}^l))\\
    % E_\mathbf{I} &= SumPooling(\phi_\mathbf{I}(\mathbf{I}))\\
    % E &= E_\mathbf{z} + E_\mathbf{I} \label{eq: our_proxy}
    E = \phi_{z} (Concat(\mathbf{z}^P, \mathbf{z}^\mathbb{P}, \mathbf{z}^L)) + SumPool(\phi_\mathbf{I}(\mathbf{I})) \label{eq: our_proxy}
\end{equation}

\label{section: finetune}

\section{Experiments}
% We first demonstrate our model's efficacy in generating novel and diverse hits for the test pocket set (section \ref{section: pocket-conditioned-generation-experiment}); We conduct a case study generalizing our model to an out-of-distribution (OOD) and relevant protein pocket for COVID-19 (section \ref{section: sars-cov-2-protease-experiment}). We further demonstrate the utility of speed and diversity on a novel virtual screening task (section \ref{section: generative-modelling-as-pre-screening}). At last, we conduct an ablation study to verify the effectiveness of pharmacophore conditioning (section \ref{section: ablations}).

% Our method do not rely on the high quality, but smaller experimental protein-ligand structure data. Instead w
\subsection{Dataset}

% predictor only require docking data, instead of 
% \paragraph{Docking score prediction dataset}
We train and evaluate \textsc{TacoGFN} on the commonly used \textbf{CrossDocked} benchmark \citep{Francoeur2020}. We first apply the splitting and processing protocol on the CrossDocked dataset to obtain the same train and test split of the 100k protein-ligand pairs as previous methods \citep{luo20213d, peng2022pocket2mol, guan2023decompdiff}. (See details in Appendix \ref{appendix: dataset}). 
%We then dock and score using Quick Vina 2.1 (QVINA) \citep{trott2010autodock, alhossary2015fast} for all protein-ligand pairs from the training split. 
We then train our docking score predictor on the training split of the protein-ligand pairs to predict their corresponding Vina Dock scores.
Since \textsc{TacoGFN} is trained on the same set of protein-ligand pairs and evaluated on the same unseen pockets, our experimental results can be directly compared against the published results of existing structure-based generative models. 

%However, we could not fairly train existing methods on the new ZINCDock-15M dataset. Instead of containing experimental-like structure data between known binders and their target, ZINCDock-15M contains random molecules docking conformation in protein pockets. Learning a distribution of random molecules with a distribution learning objective is likely unhelpful for generating high-affinity binding molecules. Our method's advantage here is that we benefit strongly from this large-scale docking data.

% Secondly, training on a dataset of this size for one epoch will take more than 6 months on a powerful A100 GPU. (See analysis in appendix). 

% , one which closely resembles practical situations, and one to benchmark directly against existing methods.
% (Mention other model cannot train a dataset of this size, because running 3 epoch will take more than 600 days on an A100 GPU) 

%Mention CrossDock
%Mention we split 
\subsection{Training}
We train one \textsc{TacoGFN} model to generate molecules conditional to any protein target structures using predicted affinity, Synthetic Accessibility (SA), and drug-likeness (QED) as the reward. We also adopt Double GFN \citep{lau2023dgfn} to improve exploration in sparse reward domains and high-dimensional states, by initializing two networks to model the action policy: an online network and a target network.

For each training trajectory, one protein pocket is randomly drawn from the CrossDocked training set first. Then, a molecule is generated through sampling a sequence of actions using our target network which models the forward action policy $P_{target}$. The reward is then computed for that molecule with respect to the protein target. The Trajectory Balance loss (Equation \ref{eq:trajbal_objective}) is used to train the online network modelling the action policy $P_{online}$. This loss has the objective for our model to generate objects with probabilities proportional to the rewards (consisting of predicted docking score, QED and SA). Periodically, the weights of the target network are updated by the weights of the online network using a delayed strategy. This strategy reduces training instability and promotes explorations of the larger chemical space. 

\subsection{Evaluation} \label{section: evaluation setting}
In all evaluations, each structure-based generative model is tasked to produce 100 molecules (ligands) for each of the 100 unseen protein pockets from the CrossDock-100k test set.

% For consistency across methods, duplicate or invalid molecules generated are thrown away. Therefore some methods may end up with lesser than 100 molecules.
% We also report the numbers of duplicated molecules, 

% As previous distributional method often generated the same molecules over and over, it was necessary to discard any redundant molecules and regenerate. We follow this approach, although it was not necessary for TacoGFN as there were nearly zero duplication in generation. 

\textbf{Evaluation metrics.}
%The goal of structure-based drug discovery is to propose novel drug candidates. Importantly, a drug candidate must simultaneously have high affinity, desirable drug-like properties, and be easily synthesizable to be considered for further (expensive) experimental validation. 
We adopt the following commonly used metrics from \citet{guan2023d} and \citet{reidenbach2024evosbdd}: (1) \textbf{Validity} is the percentage of unique generated molecules free of reconstruction errors and disconnections as determined by RDKit. 
(2) \textbf{Vina Dock} approximates the binding energy between a generated molecule and a protein pocket, where a lower docking score indicates a higher binding affinity. 
(3) \textbf{High Affinity} measures the percentage of generated molecules with higher affinity than the reference molecule.
(4) \textbf{QED} is a measure of drug-likeness, estimating a molecule's suitability as an oral drug based on its properties \citep{Bickerton2012}.
(5) \textbf{Synthetic Accessibility (SA)} estimates difficulty of synthesizing the molecule \citep{Ertl2009}. The score is normalized between 0 and 1 using the formula $(10 - SA)/9$. 
(6) \textbf{Diversity} is calculated as the average pairwise fingerprint Tanimoto distance between molecules generated for a pocket.
(7) \textbf{Success Rate} is the percentage of molecules which pass the same criteria (QED > 0.25, SA > 0.59, Vina Dock < -8.18) as in \citet{long2022zeroshot, guan2023decompdiff, zhou2024decompopt, reidenbach2024evosbdd}.
(8) \textbf{Time} is the average runtime (in seconds) for generating 100 unique and valid molecules for a pocket. 

\textbf{Baselines and problem settings.} Similar to \citet{zhou2024decompopt} and \citep{reidenbach2024evosbdd}, we separate existing methods by their problem definitions:

1) \textbf{Generative} methods are expected to generalize for pocket structures unseen during training.
%by learning from the data or reward distribution of the training protein-ligand complexes. 
They generate molecules for test pockets at inference time without optimization loops or access to docking programs. Therefore, methods under this setting are only allowed to generate 100 molecules for each pocket in one-shot. We compare \textbf{TacoGFN} trained on CrossDocked-100k against the following generative models: \textbf{liGAN} \citep{D1SC05976A}, \textbf{GraphBP} \citep{liu2022graphbp}, \textbf{AR} \citep{luo20213d}, \textbf{Pocket2Mol} \citep{peng2022pocket2mol}, \textbf{TargetDiff} \citep{guan2023d}, \textbf{DiffSBDD} \citep{schneuing2023structurebased}, \textbf{DecompDiff} \citep{guan2023decompdiff}. For consistency with existing baselines, molecules are docked using QVina in this problem setting \citep{trott2010autodock, alhossary2015fast}. For details on the docking protocol for the generative setting, please see Appendix \ref{appendix: docking details}.

2) \textbf{Optimization} methods are able to leverage docking on the target pocket. Unlike the generative setting, these methods typically iteratively optimize the candidate pool and select the top 100 molecules. We compare against the following methods: \textbf{RGA} \citep{fu2022reinforced} and \textbf{EvoSBDD} \citep{reidenbach2024evosbdd} use a black-box algorithms to conduct rounds of the optimization process, with molecule fitness based on docking score to the target pocket. \textbf{DecompOpt} and \textbf{TargetDiff+Opt} \citep{zhou2024decompopt} optimizes molecules generated by pre-trained SBDD models via rounds of optimization process involving re-docking to the target pocket. \textbf{TacoGFN+FT} fine-tunes a pre-trained \textsc{TacoGFN} to tailor the model to the target pocket using docking as the reward.

\begin{table}
\centering
\caption{Comparison of the properties of molecules generated in the \textbf{generative} problem setting for the CrossDocked test set pockets. The reference molecules are from the CrossDocked test set. The best results are in \textbf{bold}. The average and median values are calculated over the averages for each pocket. The prior method results are taken from their publication.}
\begin{adjustbox}{width=1\linewidth}
\renewcommand{\arraystretch}{1.1}
\begin{tabular}{l|c|cc|cc|cc|cc|cc|c|c}
\toprule
        & \multicolumn{1}{c|}{Validity ($\uparrow$)} & \multicolumn{2}{c|}{Vina Dock ($\downarrow$)} &
        \multicolumn{2}{c|}{High Affinity ($\uparrow$)} & 
        \multicolumn{2}{c|}{QED ($\uparrow$)}  & \multicolumn{2}{c|}{SA ($\uparrow$)} & 
        \multicolumn{2}{c|}{Diversity ($\uparrow$)} & \multicolumn{1}{c|}{Success Rate ($\uparrow$)} & \multicolumn{1}{c}{Time ($\downarrow$)} \\
        
Model  & Avg. & Avg. & Med.& Avg. & Med. & Avg. & Med. & Avg. & Med. &  Avg. & Med. & Avg. & Gen \\
\midrule
Reference & - & -7.45 & -7.26 & - & - & 0.48 & 0.47 & 0.73 & 0.74  & - & - & 25.0\% & -   \\
\midrule
liGAN & - & -6.33 & -6.20 & 21.1\% & 11.1\% & 0.39 & 0.39 & 0.59 & 0.57 & 0.66 & 0.67 & 3.9\% & - \\
GraphBP  & - & -4.80 & -4.70 &  14.2\% & 6.7\% & 0.43 & 0.45 & 0.49 & 0.48 & \textbf{0.79} & \textbf{0.78} & 0.1\% & 10 \\
AR & 92.95\% & -6.75 & -6.62 & 37.9\% & 31.0\% & 0.51 & 0.50 & 0.63 & 0.63 & 0.70 & 0.70 & 7.1\% & 19659 \\
Pocket2Mol & \underline{98.31\%} &  -7.15 & -6.79 & 48.4\% & 51.0\% & 0.56 & \underline{0.57} & \underline{0.74} & \underline{0.75} & 0.69 & 0.71 & 24.4\% & 2504  \\
TargetDiff  & 90.35\%  & -7.80 & -7.91 & 58.1\% & 59.1\% & 0.48 & 0.48 & 0.58 & 0.58 & 0.72 & 0.71 & 10.5\% & 3428  \\
DiffSBDD  & 85.01\% & -8.03 & -7.75 & 55.3\% & 56.6\% & 0.47  & 0.47 & 0.55 & 0.56 & \underline{0.76} & \underline{0.76} & 6.0\% & 160  \\
DecompDiff  & 71.96\% & \textbf{-8.39} & \underline{-8.43} & \underline{64.4\%} & \underline{71.0\%} & 0.45 & 0.43 & 0.61 & 0.60 & 0.68 & 0.68 & {24.5\%} & 6189 \\
\textsc{TacoGFN} (Ours) & \textbf{100\%} & \underline{-8.24} & \textbf{-8.44}  & \textbf{67.5\%} & \textbf{92.0\%} & \textbf{0.67} &  \textbf{0.67} & \textbf{0.79} & \textbf{0.79} & 0.53 & 0.53 & \textbf{56.0\%} & \textbf{4}  \\
\bottomrule
\end{tabular}
\renewcommand{\arraystretch}{1}
    \end{adjustbox}
    \label{tab: generative table}
\end{table}

\subsection{Experimental results} \label{section: pocket-conditioned-generation-experiment}
\begin{figure}[t]
  \centering
  \includegraphics[width=1.0\textwidth]{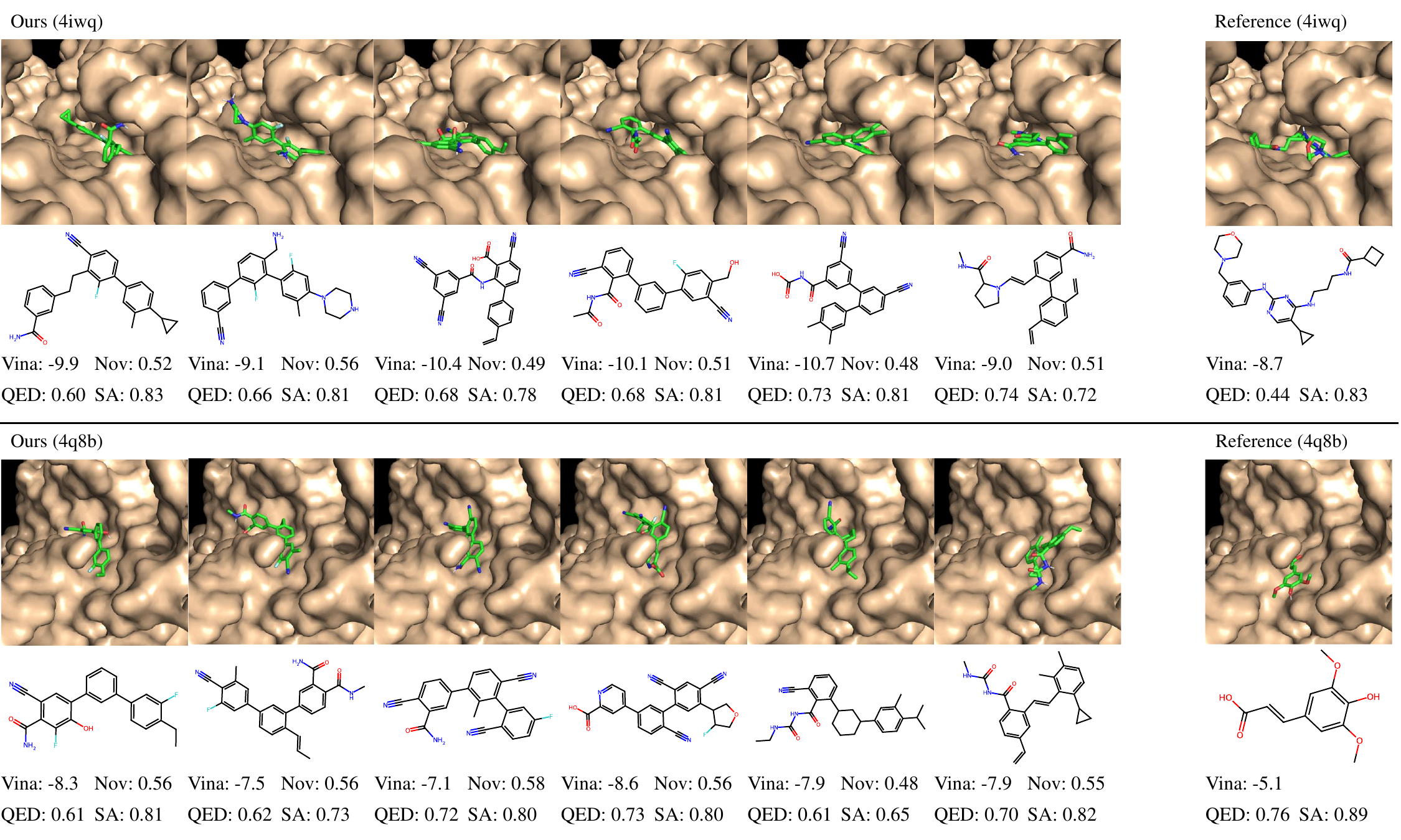}
  \caption{
    Our method is focused on de novo hit discovery - finding novel and diverse high-scoring hits for a protein target. Our method does not optimize based on a given reference seed compound. The goal of reward-based sampling is to sample diverse high-scoring molecules. To provide a fair overview of the model's performance, we selected protein pockets \textit{4iwq} and \textit{4q8b}, which are at the 25th and 75th percentiles, respectively, based on their docking scores with their native ligands. We show compare the molecules generated by \textsc{TacoGFN} against the native ligand with their QED, SA, Novelty, and Docking score (Vina). 
  }
  \label{fig: generated_molecules}
\end{figure}

\begin{figure*}[t]
  \centering
  \includegraphics[width=0.95\textwidth]{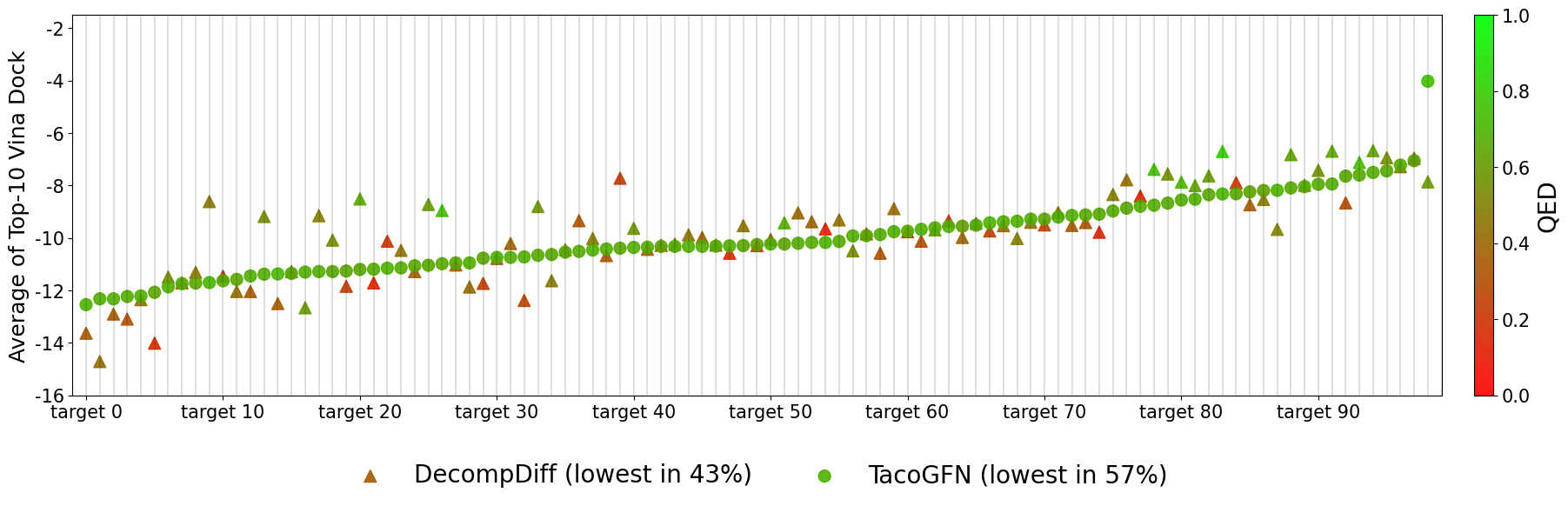}
  \caption{
    The average of the top-10 Vina Dock of molecules generated for individual CrossDocked test pockets (target) by DecompDiff and \textsc{TacoGFN}. Targets are sorted by the average of the top-10 docking score of \textsc{TacoGFN} generated molecules. A lower docking score means a higher estimated binding affinity. Color is used to denote the average QED value of molecules in the Top-10 set. A higher QED indicates the molecule is more drug-like.   
  }
  \label{fig: qed rank}
\end{figure*}

Table \ref{tab: generative table} compares \textsc{TacoGFN} against other baselines in the generative problem setting for SBDD. We examine the docking score performances for individual pockets in Figure \ref{fig: qed rank} and \ref{fig: mw rank}. We also show examples of generated molecules in Figure \ref{fig: generated_molecules} and analyze their average physical properties in Table \ref{tab: ablation validity}. We then compare \textsc{TacoGFN+FT} against existing methods for the optimization problem setting in Table \ref{tab: optimization table} and \ref{tab: optimization ranking table}, where docking is used in optimization rounds. Lastly, we conduct ablation studies to show the benefits of using a larger docking score dataset in Table \ref{tab: ablation dataset} and validate the utility of the pocket conditioning in Table \ref{tab: ablation pocket}. 

\textbf{Generative setting.} Table \ref{tab: generative table} highlights the strong performance of \textsc{TacoGFN} in the generative problem setting for the CrossDocked test set pockets. Notably, \textsc{TacoGFN} boasts a significant improvement in success rate at 56.0\%, more than doubling the previous best of 24.5\% achieved by DecompDiff. This improvement is due to \textsc{TacoGFN}'s ability to generate high-affinity molecules that simultaneously satisfy the drug-likeness and ease-of-synthesis requirements. In fact, \textsc{TacoGFN} records the best QED, SA, and High Affinity simultaneously among all \textbf{generative} methods. In Figure \ref{fig: generated_molecules}, we show examples of molecules generated by \textsc{TacoGFN} and show its ability to generate molecules with significantly improved docking scores compared to native ligands.

%  The method also excels in drug-likeness and synthesis accessibility, achieving the best QED (0.67) and SA (0.79) values.
% \textsc{TacoGFN} records the highest median Vina Dock score (-8.44). 

% In terms of affinity, \textsc{TacoGFN} demonstrates remarkable results with the highest average and median High Affinity scores (67.5\% and 92.0\%, respectively). 

It is difficult to discover molecules that simultaneously exhibit better QED and Vina Dock compared to known binders and molecules from existing baselines due to the trade-off between Vina Dock and QED. Molecules with higher molecular weight are more likely to have strong Vina Dock because of the presence of more interacting atoms, but they tend to have worse drug-like properties (QED). \footnote{This trade-off is clearly observed with Pocket2Mol and DecompDiff (see Table \ref{tab: generative table} and \ref{tab: ablation validity}), where DecompDiff has strong Vina Dock but lower QED, while Pocket2Mol exhibits the opposite trend.} In Figure \ref{fig: qed rank}, we show the performance breakdown for the 100 test protein pockets. \textsc{TacoGFN} achieves better average Top-10 Vina Dock than DecompDiff in $57\%$ of the test pockets. Notably, the top molecules generated by \textsc{TacoGFN} also consistently demonstrate higher QED values. As shown in Appendix Figure \ref{fig: mw rank}, in the pockets where DecompDiff achieves a lower Top-10 Vina Dock, the molecules often exhibit a molecular mass larger than 500 daltons. These heavier molecules violate the ideal properties of orally active drugs according to the Rule of 5 \citep{LIPINSKI19973}. In addition, heavy molecules with high docking scores are more likely to be false positives \citep{Pan2002}. \textsc{TacoGFN} is able to find molecular spaces that improve both QED and Vina Dock requirements not only by modelling the reward distribution but also by exploring a broader chemical space. This is achieved through learning from generated examples using our online policy, rather than being limited to the training data.

We note exploring millions of molecules online could not be easily achieved with existing SBDD baseline. This is because the generation process of \textsc{TacoGFN} is a few orders of magnitude faster than existing autoregressive or diffusion-based generative methods. Additionally, \textsc{TacoGFN} achieves 100\% in validity and uniqueness, demonstrating the efficiency of our fragment-based 2D generation framework. As shown in Table \ref{tab: generative table}, \textsc{TacoGFN} achieves a significant improvement in both time and validity over previous methods.   

Molecules generated from \textsc{TacoGFN} have more ideal molecular weight and drug-likeness properties, in addition to achieving better Vina Dock; Therefore, they are more suitable as drug candidates. \textsc{TacoGFN} samples at reward temperature $\beta$ of 64 at inference time, resulting in the modelling of the probability $p(x|c) \propto R(x, c)^{64}$. This policy focuses more on the modes of reward function. Therefore the molecules sampled have higher quality but lower diversity than those generated by the methods that do not attempt to satisfy multiple objectives. By changing $\beta$ or reward function we can trade off rewards with diversity as shown in Appendix \ref{appendix: diversity}.

% \textcolor{magenta}{[SEO]On the other hand, molecules generated from \textsc{TacoGFN} have more ideal molecular weight (99.76\% molecules exhibit a molecular mass between 160 480) and drug-likeness properties, in addition to achieving better Vina Dock; Therefore, they are more suitable as drug candidates.}

% As reported in Table \ref{tab: generative table}, the molecules generated by \textsc{TacoGFN} for one particular pocket have lower diversity than those generated by the methods that do not attempt to satisfy multiple objectives. 

% \textcolor{red}{}
% \newpage

\paragraph{Evaluation of physical properties.} We evaluate the following physical properties important for small molecule drugs in Table \ref{tab: ablation validity}: (1) \textbf{Ideal Mol. Wt.} is the percentage of molecules satisfying molecular weight within 160 - 480 daltons - the typical acceptable molecular weight range for small molecule drugs  \citep{ghose1999knowledge}. (2) \textbf{Mol. Wt.} is the molecular weight of molecules in Daltons. (3) \textbf{Num Heavy Atom} is the heavy atom (any atom that is not hydrogen) count of the molecules generated. (4) \textbf{Strain Energy} is the difference between the re-docked poses (meaning generated molecules are docked with RDKit's ETKDG conformation initialization), versus the strain energy of their unstrained ETKDG poses. We consider the sum of torsional, steric and electrostatic strains. If the strain energy of the ligand’s docked pose is very high, the physical compatibility for such binding becomes poor \citep{Perola2004, Gu2021}.

For baseline, we compare against the top three generative methods ranked by success rate: \textbf{DecompDiff} \cite{guan2023decompdiff}, \textbf{Pocket2Mol} \citep{peng2022pocket2mol} and \textbf{TargetDiff} \cite{guan2023d}. Additionally, we sample a random set of molecules from the ZINC virtual library for comparison. 

\begin{table*}[h]
    \centering
    \caption{ 
        Comparison on the number of heavy atoms, strain energies
    }
    \begin{tabular}{l|c|cc|cc|cc}
        \toprule
        & \multicolumn{1}{c|}{Ideal Mol. Wt. ($\uparrow$)} & \multicolumn{2}{c|}{Mol. Wt.}  & \multicolumn{2}{c|}{Num Heavy Atom} & \multicolumn{2}{c}{Strain Energy ($\downarrow$)}\\
        Model & Avg. & Avg. & Med. & Avg. & Med.  & Avg. & Med. \\
        \midrule
        ZINC molecules & 98.73\% & 337.63 & 337.34 & 23.53 & 23.50 & 377.77 & 375.17 \\
        % FDA approaved & 69.55\% & 428.45 & 344.24 & 29.59 & 24.00 & - & - \\
        \midrule
        Pocket2Mol & 64.55\% & 248.65 & 230.99 & 18.29 & 16.86 & 254.21 & 218.52 \\
        TargetDiff & 78.37\% &  323.64 & 327.31 & 22.78 & 23.15 & 600.31 & 547.86 \\
        DecompDiff & 51.18\% & 494.05 & 487.51 & 34.93 & 34.00 & 834.72 & 781.34 \\
        TacoGFN (ours) & 99.76\% & 402.56 & 402.65 & 30.47 & 30.52 & 360.51 & 357.33 \\
        \bottomrule
    \end{tabular}
    \label{tab: ablation validity}
\end{table*}

As shown in Table \ref{tab: ablation validity}, 99.76\% of molecules generated by \textsc{TacoGFN} have molecular weights within the 160 to 480 dalton range - the ideal range for small molecule drugs. This ratio is much higher than those of existing methods such as DecompDiff (51.18\%), Pocket2Mol (64.55\%) and TargetDiff (78.37\%). For Pocket2Mol, molecules fall outside of the ideal molecular weight range because they are too small (below 160 Daltons); While for DecompDiff, most molecules are outside of range because they are too heavy (greater than 480 Daltons). We note there are indeed drug candidates with molecular weights or drug-likeness outside of the typical acceptable range. However, they may be regarded as an exception rather than the norm (from a specialized drug class) or have elevated risks for poor absorption and bioavailability \citep{RITCHIE2014489}. 

PDBbind \citep{Liu2014}, the parent database of the CrossDocked set \citep{Francoeur2020}, contains a large number of common biomolecules such as ATP (molecular weight: 508 daltons) or sulphate ion (molecular weight: 96 daltons). This inherent bias in the training dataset hinders data distribution learning-based generative models from generating molecules that would be acceptable as real drugs. In contrast, \textsc{TacoGFN} learns the actual physical property distribution of drugs, resulting in an average molecular weight (402.56 daltons) of the generated molecules that closely matches that of FDA-approved drugs ($\sim$ 430 daltons).

With regards to strain energy, \textsc{TacoGFN} attains far lower strain energy than DecompDiff and TargetDiff. The strain energy of \textsc{TacoGFN} is comparable with that of ZINC molecules, as shown in Table \ref{tab: ablation validity}. Pocket2Mol achieves lower strain than ZINC and all other methods, due to the small molecular weight they generate. In summary, this analysis supports that molecules generated by \textsc{TacoGFN} show appropriate physical properties and compatibility with its binding pocket.

\begin{table}[t]
    \centering
    \caption{Comparison of the properties of molecules generated in the \textbf{optimization} problem setting for the CrossDocked test set pockets. The prior method results are taken from their publication.  The number of fine-tuning steps is 300 unless otherwise specified. }
    \resizebox{\linewidth}{!}{
        \begin{tabular}{l|c|cc|cc|cc|cc|cc|c|c}
            \toprule
            & \multicolumn{1}{c|}{Validity ($\uparrow$)} &  \multicolumn{2}{c|}{Vina Dock ($\downarrow$)} & \multicolumn{2}{c|}{High Affinity ($\uparrow$)} & \multicolumn{2}{c|}{QED ($\uparrow$)} &  \multicolumn{2}{c|}{SA ($\uparrow$)} & \multicolumn{2}{c|}{Diversity ($\uparrow$)} & \multicolumn{1}{c|}{Success Rate ($\uparrow$)} & \multicolumn{1}{c}{Time ($\downarrow$)} \\
            Model & Avg. & Avg. & Med. & Avg. & Med. & Avg. & Med. & Avg. & Med. & Avg. & Med. &  Avg. & Gen + Score. \\
            \midrule
            Reference & 100\% & -7.45 & -7.26 & - & - & 0.48 & 0.47 & 0.73 & 0.74  & - & - & 25.0\% & -   \\
            \midrule
            RGA & - & -8.01 & -8.17 & 64.4\% & 89.3\% & 0.57 & 0.57 & 0.71 & 0.73 & 0.41 & 0.41 & 46.2\% & - \\
            TargetDiff+Opt & -  & -8.30 & -8.15 & 71.5\% & 95.9\% & \textbf{0.66} & \textbf{0.68} & 0.68 & 0.67 & 0.31 & 0.30 & 25.8\% & >3728 \\ 
            DecompOpt & -  & -8.98 & -9.01 & 73.5\% & 93.3\% & 0.48 & 0.45 & 0.65 & 0.65 & 0.60 & 0.61 & 52.5\% & 9241 \\
            EvoSBDD ($\alpha$ = 1.3, 140R) & \textbf{100\%} &  {-10.27} & {-10.36} & 96.5\% & \textbf{100\%} & 0.53 & 0.52 & 0.75 & 0.77 & \textbf{0.63} & \textbf{0.63} & 78.8\% & 6300 \\
            EvoSBDD ($\alpha$ = 0, $\sigma$ = 1, 140R) & \textbf{100\%} &  -10.14 & -10.27 & 94.4\% & \textbf{100\%} & 0.59 & 0.59 & 0.77 & 0.77 & {0.62} & {0.62} & 86.4\% & 6300 \\
            \midrule
   
            TacoGFN+FT ($t_{QED}=0.4, t_{SA}=0.6$)   & \textbf{100\%} & \textbf{-10.78} & \textbf{-10.93} & \textbf{97.1\%} & \textbf{100\%} & 0.47 & 0.47 & 0.70 & 0.69 & 0.62 & 0.62 & \textbf{87.8\%} & 7750 \\
            
            TacoGFN+FT ($t_{QED}=0.5, t_{SA}=0.8$)  & \textbf{100\%} & \textbf{-10.33} & \textbf{-10.39} & \textbf{97.0\%} & \textbf{100\%} & 0.58 & 0.58 & \textbf{0.82} & \textbf{0.82} & 0.61 & 0.61 & \textbf{88.8\%} & 6230 \\
            
            TacoGFN+FT ($t_{QED}=0.6, t_{SA}=0.8$)   & \textbf{100\%} & {-10.14} & {-10.21} & 96.2\% & \textbf{100\%} & 0.65 & 0.65 & \textbf{0.82} & \textbf{0.82} & 0.59 & 0.59 & 86.3\% & {6140} \\
            \bottomrule        
        \end{tabular}
    }
    \label{tab: optimization table}
\end{table}

\paragraph{Optimization setting.} Here we study the optimization problem setting, where methods can conduct optimization rounds leveraging docking program on the target pocket. We evaluate \textsc{TacoGFN+FT}, where we fine-tune a pre-trained \textsc{TacoGFN}, using UniDock \citep{Yu2023} \footnote{Uni-Dock is a recently proposed GPU-accelerated molecular docking program that achieves more than 2000-fold speed-up compared with the AutoDock Vina running in single CPU core.}  on the target pocket as the affinity reward for generated molecules.
% We observe the docking score to further improve if it were fine-tuned for longer. 
For each protein pocket, we chose to finetune \textsc{TacoGFN} for 300 steps (unless otherwise specified) to match the time for \textsc{TacoGFN+FT} ($t_{QED}=0.5, t_{SA}=0.8$) with EvoSBDD's \citep{reidenbach2024evosbdd}.
We report the time taken for fine-tuning and docking molecules for our method, measured on a single A4000 GPU - a GPU with less performance than the one used by EvoSBDD (A6000 GPU), or TargetDiff+Opt/DecompOpt (A100 GPU). 
% This approach leverages information gained from pre-training across protein pocket structure and improves the performance through fine-tuning.
The details of the fine-tuning process and UniDock can be found in Appendix \ref{appendix: finetuning details}.

\textsc{TacoGFN+FT} is able to simultaneously optimize a set of potentially conflicting objectives - consisting of Vina Dock, QED and SA. 
% We can change $t_{QED}$ and $t{SA}$, which are threshold caps reward given to a molecules after QED or SA passes this threshold, in the reward function to priortize on different properties. 
% Our method is able to generate different molecule sets that . 
The \textsc{TacoGFN+FT} with balanced objective ($t_{QED}=0.5, t_{SA}=0.8$) simultaneously achieves state-of-art results in Vina Dock, high affinity, synthetic accessibility and success Rate. Since \textsc{TacoGFN} samples molecules during fine-tuning with a lower reward temperature $\beta$ (sampled uniformly between 0 and 64), the diversity of \textsc{TacoGFN+FT} is higher than \textsc{TacoGFN} and comparable with EvoSBDD. Our method variant ($t_{QED}=0.4, t_{SA}=0.6$) focusing on affinity attains the best average Vina Dock of -10.78. Our method variant focusing on QED achieves a comparable QED to the best baseline TargetDiff+Opt (0.65 vs 0.66), but a much better Vina Dock (-10.14 vs -8.30) and SA (0.82 vs 0.68). Please see Appendix Table \ref{tab: optimization ablation table} for full results.

% The time for scoring and generating molecules for \textsc{TacoGFN+FT} is less than EvoSBDD and DecompOpt.

Overall, our proposed fine-tuning pipeline is highly efficient, as \textsc{TacoGFN+FT} surpasses state-of-art performance while taking less time. In comparison to EvoSBDD \citep{reidenbach2024evosbdd}, which does not take pocket structure as input, \textsc{TacoGFN} is already trained to generate drug-like and synthesizable molecules with binding conditions to protein pocket structure. Therefore, finetuning \textsc{TacoGFN} generates a molecular set with a better success rate and high-affinity rate for a new protein pocket. Compared to DecompOpt \citep{zhou2024decompopt}, our method does not need to traverse the 500-1000 denoising steps to generate a molecule. This means we could search for more molecules and discover higher reward candidates in a shorter amount of time.

Table \ref{tab: optimization ranking table} compares the performance of \textsc{TacoGFN+FT} by simply generating an equal number of molecules with base \textsc{TacoGFN} and ranking the top 100 molecules without finetuning (\textsc{TacoGFN+Rank}); Here we show, although \textsc{TacoGFN+Rank} already attains competitive results, our proposed fine-tuning meaningfully further improves Vina Dock and Success Rate. 

\begin{table}[h]
    \centering
    \caption{Comparison of performance between \textsc{TacoGFN} with Finetuning (\textsc{TacoGFN+FT}) with simply ranking the same number of generated molecules from \textsc{TacoGFN} by their reward (\textsc{TacoGFN+Rank}).}
    \resizebox{\linewidth}{!}{
        \begin{tabular}{l|c|cc|cc|cc|cc|cc|c|c}
            \toprule
            & \multicolumn{1}{c|}{Validity ($\uparrow$)} &  \multicolumn{2}{c|}{Vina Dock ($\downarrow$)} & \multicolumn{2}{c|}{High Affinity ($\uparrow$)} & \multicolumn{2}{c|}{QED ($\uparrow$)} &  \multicolumn{2}{c|}{SA ($\uparrow$)} & \multicolumn{2}{c|}{Diversity ($\uparrow$)} & \multicolumn{1}{c|}{Success Rate ($\uparrow$)} & \multicolumn{1}{c}{Time ($\downarrow$)} \\
            Model & Avg. & Avg. & Med. & Avg. & Med. & Avg. & Med. & Avg. & Med. & Avg. & Med. &  Avg. & Gen + Score. \\
            \midrule
            TacoGFN+FT ($t_{QED}=0.5, t_{SA}=0.8$)   & \textbf{100\%} & \textbf{-10.33} & \textbf{-10.39} & \textbf{97.0\%} & \textbf{100\%} & 0.58 & 0.58 & \textbf{0.82} & \textbf{0.82} & \textbf{0.61} & \textbf{0.61} & \textbf{88.8\%} & {6160} \\
            TacoGFN+Rank  & \textbf{100\%} & -10.09 & -10.10 & \textbf{97.0\%} & \textbf{100\%} & \textbf{0.59} & \textbf{0.59} &  \textbf{0.82} & \textbf{0.82} & 0.60 & \textbf{0.61} & 86.4\% & 5133\\
            \bottomrule        
        \end{tabular}
    }
    \label{tab: optimization ranking table}
\end{table}

\subsection{Ablation studies}
% Our method do not S w more suited for our method 
We conduct additional ablation studies using the base \textsc{TacoGFN} under the \textbf{generative} setting to examine the effect of docking score prediction accuracy and pocket conditioning on our method. 

\textbf{Effects of using higher quality docking score predictor.} Here, we study the effect of using a more accurate docking score predictor, which is trained on a larger dataset, for reward. Since training a docking score predictor does not require high-quality protein-ligand structural data such as the CrossDock-100k set, we can introduce a second, larger dataset for docking score prediction called \textbf{ZINCDock-15M}. It consists of about 15M docking simulation data - from docking 1,000 random ZINC20 \citep{irwin2020zinc20} molecules into each of the 15,207 unique pockets from CrossDock-100k training split using QVina. We then train our pharmacophore-based docking score predictor on this larger dataset. Please see Appendix Table \ref{tab:docking_predictor_performance} for a comparison of docking score accuracy between using CrossDock-100k and ZINCDock-15M. As shown in Table \ref{tab: ablation dataset}, \textsc{TacoGFN} using the docking score predictor trained on the larger ZINCDock-15M dataset demonstrates improvements in average Vina Dock, high-affinity rate and success rate. 
% \textcolor{orange}{(if it is not required, please remove) This improvement suggests that the comprehensive docking data across general ligand molecules significantly contributes to the policy's learning effectiveness.}
This confirms it is possible to leverage the easily generated large-scale docking score data to generate more novel and higher affinity molecules. 

\begin{table}[h]
    \centering
    \caption{ 
        Evaluation of \textit{de-novo} drug design performance of \textsc{TacoGFN} using docking score predictors trained on two different docking score datasets. 
    }
    % \resizebox{\linewidth}{!}{  
    \begin{tabular}{l|c|cc|cc|cc}
        \toprule
        & Docking score & \multicolumn{2}{c|}{Vina Dock ($\downarrow$)}  & \multicolumn{2}{c|}{High Affinity($\uparrow$)} & \multicolumn{2}{c}{Success Rate($\uparrow$)} \\
        Model & dataset & Avg. & Med. & Avg. & Med. & Avg. & Med.  \\
        \midrule
        \textsc{TacoGFN} & CrossDock-100k & -8.24 & -8.44 & 67.5\% & 92.0\% & 56.0\% & 61.5\%  \\
        \textsc{TacoGFN} & ZINCDock-15M & \textbf{-8.35} &  \textbf{-8.53} & \textbf{69.5\%} & \textbf{94.5\%} & \textbf{58.3\%} & \textbf{67.5\%}  \\
        \bottomrule
    \end{tabular}
    % }
    \label{tab: ablation dataset}
\end{table}

% \newpage

\textbf{Effects of pocket conditioning.} \label{section: ablations}
To examine the effect of the proposed pocket conditioning for GFlowNet, we train a molecular generation policy unconditioned on pocket information. The docking score predictor is unchanged - meaning it still predicts a docking score for a molecule with pocket information. The results are shown in Table \ref{tab: ablation pocket}. We observe that the pocket-conditioned GFlowNet achieves higher docking scores compared to the GFlowNet without pocket conditioning. We further measure the number of non-covalent interactions of a molecule to respective pocket, by obtaining the binding pose using QVina (See Appendix \ref{appendix: pocket ablation}). We demonstrate generated molecules by pocket-conditioned GFlowNet result in more non-covalent interactions of for all categories (Hydrophobic interactions, Van der Waals contacts, Hydrogen binding) in Table \ref{tab: ablation pocket interactions}. More non-covalent interactions are indicative of the molecule having better specificity to the protein target. This ablation validates that our method is indeed leveraging pocket conditioning to learn a family of molecular distribution across different pocket structures.

\begin{table}[h]
    \centering
    \caption{ 
        Effectiveness of pocket structure conditioning evaluated using Vina Dock.
    }
    % \resizebox{\linewidth}{!}{
    \begin{tabular}{l|c|cc|cc}
        \toprule
        &  Docking score  & \multicolumn{2}{c|}{Vina Dock ($\downarrow$)}  & \multicolumn{2}{c}{Top-10 Vina Dock ($\downarrow$)} \\
        Model & dataset & Avg. & Med. & Avg. & Med. \\
        \midrule
        \textsc{TacoGFN} & ZINCDock-15M & \textbf{-8.35} &  \textbf{-8.53} &  \textbf{-9.97} & \textbf{-10.22}  \\
        w/o pocket conditioning & ZINCDock-15M  & -8.04 & -8.18 & -9.65 & -9.81  \\
        \bottomrule
    \end{tabular}
    % }
    \label{tab: ablation pocket}
\end{table}

\section{Conclusion}
In this paper, we have investigated the problem of structure-based drug design. To address the limitations of methods based on distribution learning, we have framed pocket-conditioned molecule generation as learning a multi-objective reward distribution using RL. 
%We propose \textsc{TacoGFN} - a target pocket conditioned GFlowNet for generating novel molecules with high-affinity and desired properties.
To this end, we propose \textsc{TacoGFN}, a pocket structure conditioned GFlowNet which generate drug-like and high-affinity molecules with respect to any 3D pocket structure. To the best of our knowledge, \textsc{TacoGFN} is the first RL model to address the challenging task of modelling a family of reward functions induced from all pocket structures. Our model effectively explores the greater chemical space, through generating millions of molecules using the online policy during training. We have further introduced a novel pharmacophore-based affinity predictor, where coarse-graining to the protein pocket is shown to achieve more accurate and robust predictions than existing architectures and protein representations. We finally introduce \textsc{TacoGFN+FT}, which fine-tunes the generic \textsc{TacoGFN} for a given test pocket.

Our experiments on the CrossDocked2020 benchmark have demonstrated that \textsc{TacoGFN} and \textsc{TacoGFN+FT} outperform the state-of-the-art methods in terms of Vina Dock, high affinity and success rate. This demonstrates the potential of \textsc{TacoGFN} as a powerful tool for structure-based drug discovery. In future work, we plan to validate the top generated ligands for some clinically relevant protein pockets \textit{in-vitro}, i.e. in wet-lab experiments.  

% \textcolor{blue}{Future work: combining our method with virtual screening}
% The results of \textsc{TacoGFN} have so far been validated only \textit{in-silico}. 

\subsubsection*{Broader Impact Statement}
This paper presents work whose goal is to advance machine learning methods for drug discovery. Such methods are increasingly being employed in the pharmaceutical industry since they promise to greatly speed-up the lengthy process of drug discovery and to significantly reduce its huge cost. If that promise holds, these machine-learning methods will benefit patients through better care and our society through a reduction of the economic burden of drug development.


% We thank the anonymous reviewers, as well as Emanuel Bengio, Kieran Didi, Shichong Peng and Shuman Peng for providing helpful feedback and suggestions.

\bibliography{main}
\bibliographystyle{tmlr}

\appendix
% \section{Appendix}

\section{Softwares}
In this study, we used the open-sourced code for GFlowNet \citep{bengio2021flow}, PharmacoNet \citep{seo2023pharmaconet} and GVP-GNN \citep{jing2021learning}. Our models were implemented using the Pytorch \citep{NEURIPS2019_9015} and PyTorch Geometric \citep{Fey/Lenssen/2019} libraries, which enabled efficient training and evaluation. We utilized RDKit \citep{landrum2006rdkit}, a widely-used chem-informatics library, to handle the molecular structures and compute chemical properties. 
We employed the QuickVina 2.1 (QVina) \citep{alhossary2015fast} and UniDock \citep{Yu2023} for docking, and used Openbabel \citep{o2011open} and AutoDock Tools \citep{huey2012using} to generate ready-to-dock files.

\section{Problem definition compared to existing RL baselines} \label{appendix: compare}

\textsc{TacoGFN} addresses the problem of structure-based drug design (SBDD), which aims to have one model generate high-affinity molecules conditioned on any unseen protein structure. We train this one model to generate molecules conditional to different protein target structures in the CrossDocked training set, using predicted affinity, Synthetic Accessibility (SA), and drug-likeness (QED) as the reward. Then we evaluate the trained model's performance on the 100 unseen protein target structures in the CrossDocked test set.  Molecules are generated for unseen protein structures during evaluation to measure whether SBDD models have learned generalized protein-ligand interaction patterns during training.

On the other hand, many existing optimization-based RL methods \citep{bengio2021flow, Zhavoronkov2019, Korshunova2022, Jeon2020, Goel2021, MOODlee2023exploring, reidenbach2024evosbdd} focus on the target-free problem setting - meaning the target protein structure is not used as input conditioning. RGA \cite{fu2022reinforced}, DecompOpt \cite{zhou2024decompopt} and \textsc{TacoGFN+FT} use protein structure as input, however, they still require docking oracle calls to the test protein pocket during evaluation to generate a molecule set. Therefore, they also fall within the optimization-based problem setting. These models are trained or optimized to generate molecules which optimize for predicted affinity on one protein target only and then evaluate on the same target. The goal of this evaluation is to measure the effectiveness of the algorithm in terms of generating objects which match a reward distribution. Although both methods use the commonly used metrics such as affinity, Synthetic Accessibility (SA), and drug-likeness (QED) as the reward, the goal of the evaluation and the problem difficulty is different compared to the SBDD setting.

Because protein-ligand interaction is highly specific, the distribution of molecules with high affinity will vary greatly across different possible pocket structures. The SBDD setting, which requires modelling the molecule distribution conditional to any protein pocket and generalizing to unseen protein pockets, is therefore a different and more challenging task than modelling molecule distribution for a single protein pocket.

\section{Method Details}
\subsection{Additional details of pocket conditioned GFlowNet}
\subsubsection{Molecular generation actions} \label{appendix: molecular generation action}
We follow action used in previous works on fragment-based molecular generation \citep{DBLP:journals/corr/abs-1802-04364, bengio2021flow, hamidizadeh2023semisupervised} to construct molecules fragment by fragment. We present the three types of actions available to \textsc{TacoGFN} below:

\textbf{(1) FragmentAddition}: At each step, for each fragment node $v^L_i$ in the molecular graph, we apply the same MLP over its node embedding  $\mathbf{h}^{L(N)}_i$ which produces logits over the fragment vocabulary. Each logit represents the unnormalized score for attaching fragment node $v^L_i$ to a particular new fragment node ($v^L_j$) from the vocabulary. These logits correspond to the \textit{FragmentAddition} action type.
While the \textit{FragmentAddition} specifies whether fragment node $v^L_i$ connects to fragment node $v^L_j$, it does not specify how they are connected (i.e. which atom on fragment $v^L_i$ forms a bond with which atom on fragment $v^L_j$).
\begin{align*}
    \mathbf{a}_{Add} &= MLP\left(\mathbf{h}^{L(N)}\right) \\
\end{align*}
\textbf{(2) AttachmentSpecification} determines how the fragment pairs are connected. At each step, for each directional edge in the molecular graph which connects fragment $v^L_i$  to $v^L_j$, we produce a logit over the atoms of fragment $v^L_i$. Each logit represents the unnormalized score for fragment i connecting to fragment $v^L_j$ via a single bond from a particular atom on fragment $v^L_i$, based on edge embedding $\mathbf{e}^{L(N)}_{ij}$. The molecule can only be completed when all attachment edges are specified. 
\begin{align*}
    \mathbf{a}_{Attach} &= MLP\left(\mathbf{e}^{L(N)}\right) \\
\end{align*}
\textbf{(3) StopConstruction} is a graph action that marks the finish of a molecule. The logit is produced from a single MLP output based on the final graph embedding $\mathbf{g}^{L(N)}$. All logits are concatenated and scaled into probabilities using the softmax function, and an action is sampled from the distribution. The same process is repeated for each time step until the \textit{stop construction} action is sampled. 
\begin{align*}
    \mathbf{a}_{Stop} &= MLP\left(\mathbf{g}^{L(N)}\right) \\    
\end{align*}
After all the scores for all possible actions are computed, a final action is sampled based on the equations as follows:
\begin{align*}
    P_a(\mathbf{a}|\mathcal{G}^L_t, \mathbf{h}_{\mathcal{G}^\mathcal{P}}) &= softmax\left(Concat\left(\mathbf{a}_{Add}, \mathbf{a}_{Attach}, \mathbf{a}_{Step}\right)\right) \\
    a &\sim P_a(\mathbf{a}|\mathcal{G}^L_t, \mathbf{h}_{\mathcal{G}^\mathcal{P}})    
\end{align*}

\subsubsection{Reward function.}\label{appendix: Reward function} Our reward function consists of properties highly relevant for a drug candidate: Vina Docking Score (DS), Drug Likeliness (QED) and Synthetic Accessibility (SA). Unless otherwise mentioned, $t_{DS}=-8.0$, $t_{QED}=0.7$ and $t_{SA}=0.8$ for all models. 

\paragraph{Drug Likeliness (QED) and Synthetic Accessibility (SA):} We first normalize the raw SA score using the formula $\frac{10-SA}{9}$ to obtain a reward between 0 and 1. While QED and SA need to meet a certain threshold to make a good drug molecule, optimizing these values beyond the threshold does not bring additional utility \citep{COLEY2021133}. Therefore, we clip reward $r$ for QED or SA to 1 when they achieve their respective threshold $t$. In other words, our model will not be incentivised to optimize QED/SA beyond their threshold value; Therefore more priority will be placed on optimizing the Vina Docking Score when these thresholds are reached. 
\begin{align*}
    r_{QED} &= \min\left(\frac{QED}{t_{QED}}, 1\right) \\ \\
    r_{SA} &= \min\left(\frac{SA}{t_{SA}}, 1\right)
\end{align*}

\paragraph{Vina Docking Score (DS): } 
% We first multiply the predicted docking score by $-1$ to obtain a positive affinity reward. 
% We then define a threshold $t_{DS}$ as the mean docking score of 1,000 random ZINC molecules for the given pocket, also multiplied by $-1$. This formula adjusts the reward from docking score relative to the average docking score for that particular pocket.
Here, we define the docking score threshold $t_{DS}$ to -8.0 kcal/mol, corresponding to 1$\mu \text{M}$ - an important requirement for a drug candidate. Since molecules will not be as useful if they do not surpass this affinity requirement, we scaled down the component of docking score not surpassing threshold $t_{DS}$ by 0.2. This has the effect of reducing rewards for molecules not surpassing this docking score threshold. Lastly, \citet{Pan2002} notes screening based on docking score is biased toward the selection of high molecular weight, as compound size may unfairly contribute to the energy score. We follow their suggestion of normalizing reward by the cube root of heavy atom count (HAC) - to reduce the false positives resulting from the molecular weight bias. Lastly, we multiply the whole term by -1 as our goal is to minimize the Vina Dock score.
\begin{align*}
    r_{DS} &= -\frac{(DS - t_{DS}) + 0.2 \times \max\left(DS, t_{DS}\right)}{\sqrt[3]{HAC}}
\end{align*}

% As the docking scores that indicate high affinity differ across pockets, we first compute $\overline{DS}$, which is the , also multiplied by $-1$. Next, we normalize the docking score improvement of a given molecule and pocket by subtracting $\overline{DS}$ from $\widehat{DS}$.

We obtain the final reward by multiplying the normalized rewards together:
\begin{align*}
    r = r_{DS} \times r_{QED} \times r_{SA}
\end{align*}

\paragraph{Model Details. }
We use Double GFN \citep{lau2023dgfn} to improve exploration in sparse reward domains and high-dimensional states. Our model is trained via the gradient descent method Adam \citep{kingma2017adam}. We list hyperparameters used in Table \ref{tab: hps} and compare the training time of our method in Table \ref{tab: training time}.
 
\begin{table}[h]
  \caption{Hyperparameters used for target conditional GFlowNet}
  \label{gflownet-hps}
  \centering
  \begin{tabular}{lc}
    \toprule
    Hyperparameters    & Values \\
    \midrule
    Num of training steps & $30,000$ \\
    Learning rate & $10^{-4}$ \\
    Weight decay & $10^{-8}$ \\
    Momentum & $0.9$ \\
    Adam eps & $10^{-8}$ \\
    Sampling $\tau$ & $0.99$ \\
    Learning rate $Z-$estimator & $10^{-3}$ \\
    Max nodes & $9$ \\
    Random action prob & $0.01$ \\
    Batch size & $8$ \\
    Training reward temp $\beta$ & $Uniform(0, 64)$ \\
    Inference reward temp & 64 \\
    Pocket cond dim & $128$ \\
    Transformer hidden dim & $256$ \\
    Num of transformer layers & $2$ \\
    QED threshold $t_{QED}$ & $0.7$ \\
    SA threshold $t_{SA}$ & $0.8$ \\
    \bottomrule
  \end{tabular}
  \label{tab: hps}
\end{table}

\begin{table}[h]
\centering
\caption{Comparison of training time. }
    \begin{tabular}{lcccc}
        \hline
        Model & Total Steps  & Batch Size & Total Time (hrs) & Hardware \\
        \hline
        Pocket2Mol  & 475k & 8 &72.6  & GTX A100 80GB\\
        TargetDiff & 300k & 4  &25.0 & GTX A100 80GB\\
        DecompDiff  & 300k & 4 & 41.7 & GTX A100 80GB\\
        \textsc{TacoGFN}  & 30k & 8 & 17.7 & RTX 3090 24GB\\
        \hline
    \end{tabular}
\label{tab: training time}
\end{table}

\subsection{Additional details of docking score predictor}\label{appendix: docking_score_predictor}
\paragraph{Motivation.}
When developing a docking score prediction model, two essential requirements are its speed of processing and its applicability to a variety of proteins and ligands.
Since the binding poses are computationally or experimentally expensive, previous affinity prediction models or docking score prediction models \citep{zhang2023deepbindgcn} predict energy by integrating the 1D representation vectors $\mathbf{z}^\mathcal{P}$ of the protein $\mathcal{P}$ and $\mathbf{z}^\mathcal{L}$ of ligand $\mathcal{L}$, respectively:
\begin{align}
    \mathbf{z}_p &= \phi_{\mathcal{P}}(\mathcal{P}) \\
    \mathbf{z}_l &= \phi_{\mathcal{L}}(\mathcal{L}) \\
    E &= \phi_{z} (Concat(\mathbf{z}_p, \mathbf{z}_l))
\end{align}

However, this approach can not consider the atom pairwise interactions between ligands and proteins due to global pooling, so it shows less generalizability to unseen ligands or proteins.
Compared to previous methods, MONN \citep{li2020monn} proposed a pairwise interaction map $\mathbf{I}$ between protein amino acid embeddings $\{\mathbf{h}^\mathcal{P}_i\}$ and ligand node embeddings $\{\mathbf{h}^\mathcal{L}_{j}\}$:
\begin{align}
    \mathbf{I} &= \{\mathbf{h}^\mathcal{P}_i\} \odot \{\mathbf{h}^\mathcal{L}_{j}\} \\
    E &= SumPool(\phi_\mathbf{I} (\mathbf{I}))
\end{align}

However, the use of full amino acid sequences is computationally expensive for large proteins.
Furthermore, in target-based drug design tasks that prioritize binding pocket information, affinity prediction over the entire protein can sometimes lead to incorrect energy calculations for different pockets.
Therefore, our study adopts the use of pharmacophores within the binding pocket as an alternative to considering the entire protein sequence.
This approach not only effectively captures the unique features of the protein pocket, but also simplifies its topology, thereby addressing the computational burden and the generalization issues.
Also, conceptually our docking score predictor determines whether a ligand atom corresponds to each pharmacophore node rather than whether it forms non-covalent interactions with each amino acid or atom.

In the table, we find \textsc{TacoGFN} (C-alpha) outperforms BigBind and DeepBindGCN, indicating that our model architecture (see equation 5 of our paper) which integrates both pairwise interactions and global pooling information is more effective than the other architectures. Moreover, we also observed that the pharmacophore representation significantly improved the performance over the residue graph representation. Notably, 	\textsc{TacoGFN} (Pharmacophore) trained with CrossDocked-100k shows better performance on ZINCDock-test compared to the other models trained with ZINCDock-15M. Given that the main challenge of non-structure-based bioactivity predictors is poor generalization performance to ligands not used in the training set \cite{chan2023embracing}, this result demonstrates that the pharmacophore representation of the pocket greatly improves both accuracy and generalization compared to the residue graph representation.

% \subsection{Training dataset}\label{appendix: Docking proxy dataset}
% We perform experiments on two datasets derived from CrossDock2020 \citep{Francoeur2020} - \textbf{CrossDock-100k} \citep{luo20213d, peng2022pocket2mol} and \textbf{ZINCDock-15M}.

% \paragraph{\textbf{CrossDock-100k.}}
% For the training and validation, we used \textbf{CrossDock-100k} proposed by \citep{luo20213d}.
% Crossdock-100k contains 100,000 protein-ligand pairs and 15,207 unique pockets for the training set.
% For the test, it contains 100 pockets not included in the training set and one ligand per pocket, in other words, it contains 100 protein-ligand pairs.

% \paragraph{\textbf{ZINCDock-15M.}}
% However, the active learning process of GFlowNet can explore a large chemical space.
% Since the ligand space of CrossDock-100k is a small fraction of the total compound space, training with only the CrossDock-100k dataset makes the docking proxy model susceptible to memorization or over-fitting problems \citep{wallach2018most, chan2023embracing}.
% Therefore, in this study, we use docking simulation dataset \textbf{ZINCDock-15M} which contains 1,000 ZINC20 \citep{irwin2020zinc20} molecules randomly selected for each of the pockets for each training dataset and test set.

\paragraph{Training details}
For model training, we use 15\% of pocket in the training set for the validation set.
We used AdamW \citep{loshchilov2018decoupled} optimizer with betas of (0.9, 0.999) and a weight decay of 0.05, and a learning rate is 0.0001 with a decaying factor of 0.1 per 20,000 iterations.
Each iteration has 32 randomly sampled pockets and 128 randomly selected ligands for each pocket during training on ZINCDock-15M.
For CrossDock-100k, we used up to 4 ligands per pocket.
We optimize our model with SmoothL1Loss \citep{girshick2015fast} for 100,000 iterations and select the best model weights with the lowest validation loss.
The training process takes about a day on 4 NVIDIA RTX A4000 GPUs.

\section{Additional experiment details}
\subsection{Dataset} \label{appendix: dataset}
CrossDocked \citep{Francoeur2020} is a dataset containing 22.5 million poses of ligands docked into multiple similar binding pockets from PDBBind. For \textbf{CrossDocked-100k} set, we use an identical processing strategy to previous works \citep{luo20213d, peng2022pocket2mol}: First, data points with binding pose RMSD greater than 1\AA \ are filtered.
Then, mmseq2 \citep{Steinegger2017} is to cluster data at 30\% sequence identity.
After splitting, 100,000 protein-ligand pairs are randomly drawn for the training set.
100 test pockets are drawn from the remaining pocket clusters.
15,307 unique protein pockets remain in the training set.

\subsection{Docking protocols} \label{appendix: docking details}
% In some previous evaluation protocols, the generated 3D molecules from the model are directly used as the conformations for the re-docking process.
% However, these generated 3D conformations may contain unrealistic bond length or bond angles \citep{harris2023benchmarking} and therefore produce distorted docking scores.
% It is more reasonable and realistic to first calculate a conformer with realistic geometry, which is a necessary step in standard docking protocols, before docking \citep{Bender2021}.
For our protocol under the generative setting, we first convert all generated molecules into SMILES and calculate their ETKDG conformers (srETKDGv3) using RDKit.
Then, we prepare ready-to-dock files of ligands and proteins with Openbabel and AutoDockTools \citep{o2011open, huey2012using}.
Finally, we dock these conformers using Quick VINA 2.1 (QVina) \citep{trott2010autodock, alhossary2015fast}. 
For QVina, we use a box size of 20~\AA\ and an exhaustiveness of 8. 
Note for the optimization setting, we adopt UniDock instead following previous works \citep{reidenbach2024evosbdd} (Please see details in Appendix \ref{appendix: unidock setting}). 

% \paragraph{Experimental metric details}
% Some baseline methods did not generate at least 10 molecules which satisfy the \textit{desirable} criteria for a number of pockets.
% To compute \textbf{Top-10 Desr. DS} in these cases, a docking score of 0 is used as a placeholder for the top 10 docking scores.   

\subsection{Baseline evaluations} \label{appendix: baseline details}

For our baseline generation, we selected Pocket2Mol \citep{peng2022pocket2mol}, TargetDiff \citep{guan2023d}, DecompDiff \citep{guan2023decompdiff}. We adhered to the default hyperparameter settings for all models. DecompDiff incorporates three prior modes: subpocket, reference, and beta. Of these, we selected the beta mode for our analysis, as it demonstrated the highest docking score, aligning with findings reported in its original publication.

We generated 100 molecules for each of the 100 protein pockets from the Crossdocked2020 \citep{Francoeur2020} test set. Following generation, we applied a filtering process to ensure that all molecules were unique and met validity criteria. Molecules were deemed invalid and discarded, if they had reconstruction errors, were duplicates, or were disconnected. The remaining SMILES obtained from these models were compiled to establish the baseline. We then proceeded to perform the docking protocol on each of the SMILES to obtain their respective docking scores. Our results were in line with those reported in the original studies of the respective models.

\subsection{Additional Results}

\subsubsection{Additional ablation on effects of pocket conditioning} \label{appendix: pocket ablation}

To further demonstrate the effectiveness of pocket conditioning, we compare the numbers of non-covalent interactions achieved by the generated molecules from pocket-conditioned TacoGFN with TacoGFN without pocket conditioning. The non-covalent interactions between protein and ligand fall into the following categories:
(1) \textbf{Hydrophobic Interactions} are interactions between nonpolar regions of the protein and the ligand.
(2) \textbf{Van der Waals Contacts} are weak forces arising from induced electrical interactions between closely positioned atoms of the protein and ligand.
(3) \textbf{Hydrogen Bonding} refers to strong dipole-dipole interactions between an electronegative atom and (typically) a hydrogen atom.
(4) \textbf{Total Interactions} refers to the total number of interactions between the protein and the ligand, including all types of non-covalent interactions.
The generated molecules are docked using QVina \citep{trott2010autodock, alhossary2015fast} to generate a binding conformation. We use PoseCheck \citep{harris2023benchmarking} to compute the number of interactions.

In table \ref{tab: ablation pocket interactions}, we found that pocket conditioning increased all types of non-covalent interactions between protein and generated molecules. this result suggests that the pocket structural information can provide guidance in selecting functional groups that can form better supramolecular interactions with the target pocket.

\begin{table}[h]
    \centering
    \caption{ 
        Effectiveness of pocket structure conditioning measured using the average numbers of non-covalent interactions (NCI) achieved after docking the generated molecules into the target pocket using QVina. Higher numbers of interactions indicate a stronger binding interaction. 
    }
    \resizebox{\linewidth}{!}{
    \begin{tabular}{l|ccccc}
        \toprule
        Model &  Docking score dataset  & Hydrophobic  ($\uparrow$) &  Van der Waals contacts  ($\uparrow$) & Hydrogen bonding ($\uparrow$) & Total interactions ($\uparrow$)  \\
        \midrule
        \textsc{TacoGFN} & ZINCDock-15M &  \textbf{5.67} &  \textbf{7.23} & \textbf{1.15} & \textbf{14.05}  \\
        w/o pocket conditioning & ZINCDock-15M   & 5.19 & 6.83 & 1.11  & 13.14 \\
        \bottomrule
    \end{tabular}
    }
    \label{tab: ablation pocket interactions}
\end{table}

\subsubsection{Discussion on diversity} \label{appendix: diversity}
In our experiments with GFlowNet, we find a trade-off between diversity and the goal of optimizing for a higher average affinity of the generated candidate set. As \textsc{TacoGFN} seeks to generate molecules with a strong affinity to a specific pocket structure while optimizing for QED and SA, the resulting solution set is inevitably smaller, leading to reduced diversity.

We note that the molecular diversity of \textsc{TacoGFN}, if desired, can be increased in various ways. Firstly, the reward is currently exponentiated by reward temperture $\beta$. As $\beta$ increases, the reward density is concentrated close to the modes with high reward, and subsequently, there’s a reduction in diversity \citep{jain2023multiobjective}. Diversity can be improved by lowering the $\beta$.

\begin{table}[h]
\centering
\caption{Comparison of average Vina Dock and diversity of \textsc{TacoGFN} trained under different reward settings. }
    \begin{tabular}{lcccc}
        \hline
        Model & Vina Dock ($\downarrow$) & Diversity ($\uparrow$) \\
        \hline
        \textsc{TacoGFN} & \textbf{-8.24} & 0.53 \\
        $\textsc{TacoGFN}_{lenient}$ & -7.91 & \textbf{0.73}  \\
        \hline
    \end{tabular}
\label{tab: training time}
\end{table}

In addition, we have experimented with various reward function settings. In the reward scenario adopted in the paper - the strict setting (see Appendix \ref{appendix: Reward function}), we reduce the reward given to the component docking score that is below -8.0 kcal/mol. We have found that the stricter reward produces better affinity metrics at the expense of lower diversity. In the lenient setting, where we simply reward docking scores based on their normalized raw values, $\textsc{TacoGFN}_{lenient}$ still outperforms Pocket2Mol and TargetDiff in terms of average docking score (-7.91) while having higher diversity (0.73).

In summary, if the key concern is candidate diversity and novelty, \textsc{TacoGFN} can be trained to generate generally high reward molecules under the lenient setting and/or low $\beta$. If the goal is to generate the best compound set with high average affinity and top affinity with decent diversity, and satisfying QED and SA, \textsc{TacoGFN} can be rewarded under the strict setting and/or high $\beta$.

\begin{figure*}[t]
  \centering
  \includegraphics[width=0.95\textwidth]{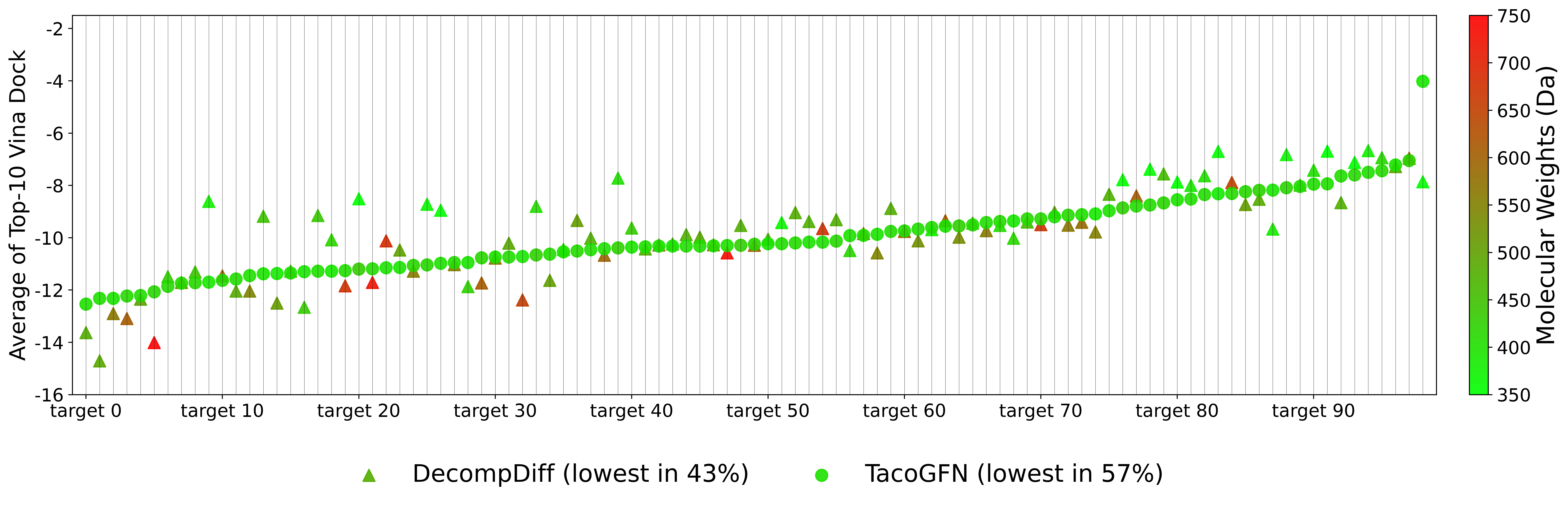}
  \caption{
    The average of the top-10 Vina Dock of molecules generated for individual CrossDocked test pockets (target) by DecompDiff and \textsc{TacoGFN}. Color is used to denote the average molecular weight of molecules in the Top-10 set. Molecular mass of an orally active drug should be less than 500 daltons \cite{LIPINSKI19973}; Heavy molecules with high docking scores are more likely to be false positives \citep{Pan2002}. Overall 	\textsc{TacoGFN} consistently achieves ideal molecular weight and strong Vina Dock. 
  }
  \label{fig: mw rank}
\end{figure*}

\subsubsection{Performance of docking score prediction model}

While many existing affinity-prediction architectures either work for one target only \citep{bengio2021flow, gentile2020deepdocking}, or require binding complex structures \citep{shen2022rtmscore}, we cannot adopt these model architectures as we require affinity prediction across various protein targets without using their binding complex structure. Thus, we compare our pharmacophore-based affinity prediction architecture with BigBind \citep{Brocidiacono2022} and DeepBindGCN \citep{zhang2023deepbindgcn}, two recently proposed non-complex-based methods which achieved state-of-the-art performance in virtual screening and affinity prediction, respectively. In addition, we examine the effect of introducing pharmacophore-based coarse-grained encoding of the protein pocket, by creating an ablation where we use residue graphs with C-alpha as nodes (as typically done in other methods) \cite{lu2022tankbind} instead of pharmacophore graphs to represent our protein pocket. We call this ablation model \textsc{TacoGFN} (C-alpha). We re-train these models using the same CrossDocked2020 and ZINCDock-15M datasets and evaluate them on both test sets.

\begin{table}[h]
    \centering
    \caption{The evaluation of docking score prediction performance of predictor trained on two different docking score datasets.}
    \begin{tabular}{lccccc}
        \toprule
        Model & Training set  & \multicolumn{2}{c}{CrossDocked-100k-test}  & \multicolumn{2}{c}{ZincDock-15M-test} \\
        & & RMSE & MAE & RMSE & MAE \\
        \midrule
        \multirow{2}{*}{BigBind} & CrossDocked-100k & 1.252            & 0.991           & 1.516 & 1.175    \\           
        & ZINCDock-15M                    & 1.561           & 1.227           & 1.233    & 0.955    \\
        \midrule
        \multirow{2}{*}{DeepBindGCN} & CrossDocked-100k & 	1.739& 1.366 & 1.912&1.531    \\           
        & ZINCDock-15M                    & 1.877& 1.470	& 1.409&	1.066   \\
        \midrule
        \multirow{2}{*}{\textsc{TacoGFN} (C-alpha)} & CrossDocked-100k & 1.091 &        0.821           & 1.517 & 1.157    \\           
        & ZINCDock-15M       & 1.483            & 1.156           & 1.170    & 0.872    \\
        \midrule
        \multirow{2}{*}{\textsc{TacoGFN} (Pharmacophore)} & CrossDocked-100k & \textbf{0.881}            & \textbf{0.652}           & \textbf{1.121} & \textbf{0.791}    \\           
        & ZINCDock-15M                    & \textbf{1.143}           & \textbf{0.880}           & \textbf{0.862}    & \textbf{0.574}    \\
        \bottomrule
    \end{tabular}
    \label{tab:docking_predictor_performance}
\end{table}

To evaluate the performance of our docking score predictor, we use two test sets: \textbf{CrossDocked-100k-test} and \textbf{ZINCDock-test}.
\textbf{CrossDocked-100k-test} is the test set of CrossDocked-100k dataset used in \citep{luo20213d} and contains one ligand per each pocket.
We re-docked each test ligand to perform the evaluation in the same environment as the generation process.
\textbf{ZINCDock-test} is the docking simulation data of randomly selected 100 ZINC20 molecules for each pocket in CrossDocked-100k-test.
The size of the test sets is 100 pocket-ligand pairs for the CrossDocked-100k-test and 10,000 pocket-ligand pairs for the ZINCDock-test.

\section{Finetuning of TacoGFN} \label{appendix: finetuning details}

\begin{table}[h]
    \centering
    \caption{\textsc{TacoGFN+FT} ablations: Here, we vary the reward function parameters. }
    \resizebox{\linewidth}{!}{
        \begin{tabular}{l|c|cc|cc|cc|cc|cc|c|c}
            \toprule
            & \multicolumn{1}{c|}{Validity ($\uparrow$)} &  \multicolumn{2}{c|}{Vina Dock ($\downarrow$)} & \multicolumn{2}{c|}{High Affinity ($\uparrow$)} & \multicolumn{2}{c|}{QED ($\uparrow$)} &  \multicolumn{2}{c|}{SA ($\uparrow$)} & \multicolumn{2}{c|}{Diversity ($\uparrow$)} & \multicolumn{1}{c|}{Success Rate ($\uparrow$)} & \multicolumn{1}{c}{Time ($\downarrow$)} \\
            Model & Avg. & Avg. & Med. & Avg. & Med. & Avg. & Med. & Avg. & Med. & Avg. & Med. &  Avg. & Gen + Score. \\
            \midrule
            Reference & 100\% & -7.45 & -7.26 & - & - & 0.48 & 0.47 & 0.73 & 0.74  & - & - & 25.0\% & -   \\
            \midrule
            RGA & - & -8.01 & -8.17 & 64.4\% & 89.3\% & 0.57 & 0.57 & 0.71 & 0.73 & 0.41 & 0.41 & 46.2\% & - \\
            TargetDiff+Opt & -  & -8.30 & -8.15 & 71.5\% & 95.9\% & \textbf{0.66} & \textbf{0.68} & 0.68 & 0.67 & 0.31 & 0.30 & 25.8\% & >3728 \\ 
            DecompOpt & -  & -8.98 & -9.01 & 73.5\% & 93.3\% & 0.48 & 0.45 & 0.65 & 0.65 & 0.60 & 0.61 & 52.5\% & 9241 \\
            EvoSBDD ($\alpha$ = 1.3, 140R) & \textbf{100\%} &  {-10.27} & {-10.36} & 96.5\% & \textbf{100\%} & 0.53 & 0.52 & 0.75 & 0.77 & \textbf{0.63} & \textbf{0.63} & 78.8\% & 6300 \\
            EvoSBDD ($\alpha$ = 0, $\sigma$ = 1, 140R) & \textbf{100\%} &  -10.14 & -10.27 & 94.4\% & \textbf{100\%} & 0.59 & 0.59 & 0.77 & 0.77 & {0.62} & {0.62} & 86.4\% & 6300 \\
            \midrule
            % TacoGFN+FT 0.4,0.6 (Ours)  & \textbf{100\%} & \textbf{-10.35} & \textbf{-10.54} & \textbf{96.6\%} & \textbf{100\%} & 0.53 & 0.52 & \textbf{0.80} & \textbf{0.80} & 0.61 & 0.61 & \textbf{87.7\%} & \{5980} \\
            % TacoGFN+FT 0.5,0.8 (Ours)  & \textbf{100\%} & -\{10.24} & \textbf{-10.39} & \{95.9\%} & \textbf{100\%} & 0.58 & 0.58 & \textbf{0.82} & \textbf{0.83} & 0.60 & 0.60 & \{86.2\%} & \{5980} \\
            % 0.4, 0.6, 1.5, Balance

            TacoGFN+FT ($t_{QED}=0.40, t_{SA}=0.60, n=240$)   & \textbf{100\%} & {-10.63} & {-10.78} & \textbf{97.1\%} & \textbf{100\%} & 0.48 & 0.47 & 0.72 & 0.71 & 0.62 & 0.62 & 86.5\% & {6200} \\

            TacoGFN+FT ($t_{QED}=0.50, t_{SA}=0.80, n=240$)   & \textbf{100\%} & -10.24 & {-10.31} & 97.0\% & \textbf{100\%} & 0.58 & 0.58 & 0.82 & 0.82 & 0.61 & 0.61 & 87.7\% & {4980} \\

            TacoGFN+FT ($t_{QED}=0.50, t_{SA}=0.75, n=240$)   & \textbf{100\%} & -10.21 & {-10.31} & 96.2\% & \textbf{100\%} & 0.58 & 0.57 & 0.80 & 0.81 & 0.62 & 0.61 & 86.9\% & {5210} \\

            TacoGFN+FT ($t_{QED}=0.55, t_{SA}=0.80, n=240$)   & \textbf{100\%} & -10.16 & {-10.27} & 96.8\% & \textbf{100\%} & 0.62 & 0.61 & 0.82 & 0.82 & 0.60 & 0.60 & 86.8\% & {5010} \\

            TacoGFN+FT ($t_{QED}=0.60, t_{SA}=0.80, n=240$)   & \textbf{100\%} & -10.07 & {-10.11} & 95.8\% & \textbf{100\%} & 0.65 & 0.64 & 0.82 & 0.82 & 0.60 & 0.60 & 85.7\% & {4910} \\

            \midrule
            
            TacoGFN+FT ($t_{QED}=0.40, t_{SA}=0.60, n=300$)   & \textbf{100\%} & \textbf{-10.78} & \textbf{-10.93} & \textbf{97.1\%} & \textbf{100\%} & 0.47 & 0.47 & 0.70 & 0.69 & 0.62 & 0.62 & 87.8\% & {7750} \\
            
            TacoGFN+FT ($t_{QED}=0.50, t_{SA}=0.80, n=300$)   & \textbf{100\%} & {-10.32} & {-10.39} & 97.0\% & \textbf{100\%} & 0.58 & 0.58 & 0.82 & 0.82 & 0.61 & 0.61 & \textbf{88.8\%} & {6230} \\
            
            TacoGFN+FT ($t_{QED}=0.50, t_{SA}=0.75, n=300$)   & \textbf{100\%} & {-10.31} & {-10.41} & 96.2\% & \textbf{100\%} & 0.57 & 0.57 & 0.80 & 0.80 & 0.62 & 0.62 & 87.7\% & {6520} \\
            
            TacoGFN+FT ($t_{QED}=0.55, t_{SA}=0.80, n=300$)   & \textbf{100\%} & {-10.24} & {-10.35} & 97.0\% & \textbf{100\%} & 0.62 & 0.61 & \textbf{0.83} & \textbf{0.82} & 0.60 & 0.60 & 87.3\% & {6270} \\
            
            TacoGFN+FT ($t_{QED}=0.60, t_{SA}=0.80, n=300$)   & \textbf{100\%} & {-10.14} & {-10.21} & 96.2\% & \textbf{100\%} & 0.65 & 0.65 & 0.82 & 0.82 & 0.59 & 0.59 & 86.3\% & {6140} \\
            
            \bottomrule
        \end{tabular}
    }
    \label{tab: optimization ablation table}
\end{table}

\subsection{Finetuning settings. }  \label{appendix: unidock setting} 

For finetuning \textsc{TacoGFN}, we adopt the same reward function as described in appendix section \ref{appendix: Reward function}. However, instead of using the previously mentioned docking score predictor for affinity reward, we use the docking program directly. Following the same setting as EvoSBDD \citet{reidenbach2024evosbdd}, we adopt UniDock \citep{Yu2023} - a GPU-accelerated docking program to compute docking score between the generated molecule and the protein target during the fine-tuning stage. We use UniDock's default balanced mode, which has an exhaustiveness of 384 and a max step of 40. We fine-tune using a batch size of 64 to leverage the speedup from UniDock's parallelization for computing docking scores. 

For a fair comparison with the previous methods, we re-scored the UniDock's docking pose with Vina (in a score-only mode without any structural changes) and obtained the same values within the margin of error ($\pm$ 0.005 kcal/mol).
% For the final evaluation, the poses from UniDock computed during the fine-tuning stage are re-scored with Vina (in score-only mode, without structure change) to ensure a fair comparison with previous methods \citep{reidenbach2024evosbdd}.
The top 100 molecules with the highest rewards from the fine-tuning stage are used for evaluation. 

% \subsubsection{Finetuning efficency} \label{appendix: finetuning efficency}
% For each protein pocket,  to match the EvoSBDD's \citep{reidenbach2024evosbdd} time. We observe the docking score to further improve if it were fine-tuned for longer. Our proposed fine-tuning process is highly efficient, as we surpass state-of-art performance while taking lesser time than comparable methods on a single A4000 GPU. We also leverage the speedup from UniDock's parallelization for computing docking scores, due to the use of a larger batch size. Unlike EvoSBDD, which do not take pocket structure as input, TacoGFN is already trained to generate drug-like and synthesizable molecules with binding condition to protein pocket structure. Therefore, finetuning TacoGFN generates molecule set with better success rate and high affinity rate for a new protein pocket. Compared to DecompOpt, our method do not need to traverse the 500-1000 reverse diffusion steps to generate a molecule. This means we could search more molecules and discover higher reward candidates in a shorter amount of time. 

% \begin{table}[h]
%   \caption{Default hyperparameters used for fine-tuning TacoGFN}
%   \label{gflownet-hps}
%   \centering
%   \begin{tabular}{lc}
%     \toprule
%     Hyperparameters    & Values \\
%     \midrule
%     Num of training steps - $s$ & $300$ \\
%     Batch size & $64$ \\
%     QED threshold $t_{QED}$ & $0.7$ \\
%     SA threshold $t_{SA}$ & $0.8$ \\
%     \bottomrule
%   \end{tabular}
%   \label{tab: hps}
% \end{table}

\subsection{Results. }
For ablations studies, we vary our reward function by adjusting $t_{QED}$ and $t_{SA}$. Once the threshold for $t_{QED}$ or $t_{SA}$ is reached for molecules, the model will not be incentivised to optimize its QED or SA further. With a lower threshold standard ($t_{QED}$ and $t_{SA}$), our model will prioritize optimizing for Vina Dock instead of optimizing for QED or SA. (See details in Appendix \ref{appendix: Reward function}). We observe such a trade-off between QED and SA with Vina Dock in our ablation studies. We also experiment with various fine-tuning steps $n$. We show that Vina Dock metric will continue to improve with more fine-tuning steps.

\end{document}

%% file: math_commands.tex
\usepackage{amsmath,amsfonts,bm}

\def\eqref#1{equation~\ref{#1}}
\def\1{\bm{1}}

\DeclareMathAlphabet{\mathsfit}{\encodingdefault}{\sfdefault}{m}{sl}
\SetMathAlphabet{\mathsfit}{bold}{\encodingdefault}{\sfdefault}{bx}{n}